\documentclass[]{article}

\usepackage{url}
\usepackage{authblk}
\usepackage{lineno}
\usepackage{epsfig}
\usepackage{psfrag}
\usepackage{amsmath,amssymb,amsthm,mathrsfs,amsfonts,dsfont}
\usepackage{epstopdf}
\usepackage{algorithm}
\usepackage{algpseudocode}
\usepackage{subfigure}
\usepackage{multirow}
\usepackage{tikz}
\usepackage{fancyvrb}

\DefineVerbatimEnvironment%
{Sinput}{Verbatim}
{fontsize=\scriptsize}

\DefineVerbatimEnvironment%
{Soutput}{Verbatim}
{fontsize=\scriptsize}

\theoremstyle{definition}

\modulolinenumbers[5]

\newcommand{\norm}[1]{\left\lVert#1\right\rVert}
\newcommand{\mbf}[1]{\mathbf{#1}}
\newcommand{\mbs}[1]{\boldsymbol{#1}}
\newcommand{\citep}[1]{\cite{#1}}
\newcommand{\pkg}[1]{\texttt{#1}}
\newcommand{\proglang}[1]{\texttt{#1}}
\newcommand{\code}[1]{\texttt{#1}}

\title{The \texttt{MOEADr} Package -- A Component-Based Framework for 
  Multiobjective Evolutionary Algorithms Based on Decomposition\footnote{Submitted for publication in the Journal of Statistical Software on October 10, 2017.}}
\author[1]{Felipe Campelo}
\author[1]{Lucas S. Batista}
\author[2]{Claus Aranha}
\affil[1]{Operations Research and Complex Systems Laboratory\\
  Department of Electrical Engineering\\
  Universidade Federal de Minas Gerais\\
  Belo Horizonte 31270-010, Brazil\\
  E-mail: \{fcampelo,lusoba\}@ufmg.br}
\affil[2]{Department of Computer Sciences\\
  Faculty of Systems Information Engineering\\
  University of Tsukuba\\
  Tennodai 1-1-1, Tsukuba City, Ibaraki, Japan, 305-8550\\
  E-mail: caranha@cs.tsukuba.ac.jp}

\begin{document}
\date{October 10, 2017}
\maketitle

\begin{abstract}
Multiobjective Evolutionary Algorithms based on Decomposition (MOEA/D)
represent a widely used class of population-based metaheuristics for
the solution of multicriteria optimization problems. We introduce the
\texttt{MOEADr} package, which offers many of these variants as instantiations of a
component-oriented framework. This approach contributes for easier
 reproducibility of existing MOEA/D variants from the literature, as
 well as for faster development and testing of new composite
 algorithms. The package offers an standardized, modular
 implementation of MOEA/D based on this framework, which was designed
 aiming at providing researchers and practitioners with a standard way
 to discuss and express MOEA/D variants. In this paper we introduce
 the design principles behind the MOEADr package, as well as its
 current components. Three case studies are provided to illustrate the
 main aspects of the package.

\end{abstract}

\section{Introduction}
\label{sec:intro}

Multiobjective Optimization Problems (MOPs) \cite{book.Miettinen1999}
are problems in which multiple objective functions must be
simultaneously optimized by the same set of parameters . MOPs are
often characterized by a set of conflicting objective functions, which
results in the existence of a set of optimal compromise solutions,
instead of a single globally optimal one. In this way, multiobjective
optimization algorithms are frequently characterized by their ability
to find (representative samples of) this set of solutions with
different compromises between the objective functions.

Multiobjective Evolutionary Algorithms based on Decomposition
(MOEA/Ds), originally proposed by \cite{journal.Zhang2007}, represent
a widely used class of population-based metaheuristics for solving
MOPs \cite{journal.Trivedi2016}. MOEA/Ds approach the problem by
decomposing it into a number of single-objective subproblems, which
are then solved in parallel by a set of candidate solutions commonly
referred to as a \textit{population}.

Based on this general concept, a number of variations and improvements
have been proposed over the past decade
\cite{journal.Trivedi2016}. As is common in the general literature on
evolutionary algorithms, many of these improvements are presented as
monolithic entities, in which a fixed composition of operators and
adjustments to the original algorithm is presented as a single, novel
approach. Opposed to this monolithic approach,
\cite{inproc.Bezerra2015} have argued towards a more modular design of
evolutionary algorithms in general, where an optimizer is seen as a
composition of multiple, specialized components. This
\emph{component-based} approach allows researchers to clearly identify
the contribution of each component, and facilitates the automated
generation of new variants based on existing components, e.g., for the
solution of specific problem classes.  This approach also allows users
to more easily test and implement new components, streamlining the
development of new ideas and the reproducibility of results.

In this context, we have developed a component-oriented framework for
MOEA/Ds, in which modules can be easily added, removed, modified or
recombined by either users or automated testing and tuning programs.
In particular, we defined a \emph{Variation Stack}, which allows a
very flexible way to model many different combinations and use cases
of variation, local search, and solution repair operators.

This framework is implemented as the \texttt{MOEADr} package, which
contains not only the original MOEA/D components, but several
components found in more recent variations, all of which were recast
as instantiations of the proposed component-based framework. The
package is implemented in the \texttt{R} language and environment
for statistical computing, and has been published on the Comprehensive
R Archive Network (CRAN), at
\url{https://CRAN.R-project.org/package=MOEADr}. A development version
is also available at the project repository,
\url{http://github.com/fcampelo/MOEADr}.

This paper describes the package and its underlying framework, and
introduces the concepts necessary for a user to sucessfully apply
\texttt{MOEADr} to their research or application problem. First, we
provide a short primer on multiobjective optimization, as well as a
short review of the MOEA/D (Section~\ref{sec:mop}). We then review
existing implementations of the MOEA/D, as well as existing
implementations of MOP solvers in the \texttt{R} ecosystem, and
locate our contribution in this context (Section~\ref{sec:moead-rev}).

Following that, we detail the framework of the \texttt{MOEADr} package,
and present the MOEA/D components that it currently implements
(Sections~\ref{sec:moea-d} and~\ref{sec:components}). This serves as a
reference to the main features of the package, as well as a starting
point for future contributions.

Finally, in Section~\ref{sec:experiments} we describe the basic usage
of the package with three case studies: A basic example of solving a
benchmark MOP using a traditional MOEA/D algorithm; an example of
using the \texttt{MOEADr} framework in conjunction with an automated
algorithm assembling/tuning method to generate a new algorithmic
configuration for a specific problem class; and an example of adding a
custom component to the framework. Readers who are not concerned with
the theoretical underpinnings of the package can skip straight to this
section.

\section{Multiobjective optimization problems and the MOEA/D} 
\label{sec:mop}

With no loss of generality, we define a continuous MOP, subject to inequality and equality
constraints, as
\begin{equation}
\begin{split}
    &\min\limits_{\mbf{x}}~ \mbf{f}\left(\mbf{x}\right) =
\left(f_1(\mbf{x}),\dotsc,f_{n_f}(\mbf{x})\right)\\
&\mbox{ subject
	to: } \mbf{x} \in \Omega,
\end{split}
  \label{eq:mop}
\end{equation}

\noindent where $n_f$ is the number of objective functions, 
$\mbf{x} \in \mathbb{R}^{n_v}$ represents a candidate solution, 
$n_v$ is the number of decision variables,
$\mbf{f}(\cdot):\mathbb{R}^{n_v}\mapsto\mathbb{R}^{n_f}$ is a vector
of objective functions, and $\Omega\subset\mathbb{R}^{n_v}$ is the
feasible decision space, such that
\begin{equation}
  \Omega = \{ \mbf{x} \in \mathbb{R}^{n_v} \mid g_i(\mbf{x}) \leq 0
  ~\forall i \wedge h_j(\mbf{x}) = 0 ~\forall j \},
\end{equation}

\noindent where $g_i(\cdot):
\mathbb{R}^{n_v}\mapsto\mathbb{R},~i=1,\dotsc,n_g$ and
$h_j(\cdot):\mathbb{R}^{n_v}\mapsto\mathbb{R},~j=1,\dotsc,n_h$
represent the inequality and equality constraint functions,
respectively.  The image of the set $\Omega$, denoted by $\mbf{f}(\Omega)$,
defines the set of attainable objective values.


Given two feasible solutions $\mbf{x}_i,~\mbf{x}_j\in \Omega$, we say
that $\mbf{x}_i$ \textit{Pareto-dominates} $\mbf{x}_j$ (written as
$\mbf{f}(\mbf{x}_i) \prec \mbf{f}(\mbf{x}_j)$ or, equivalently, $\mbf{x}_i \prec \mbf{x}_j$ ) iif
$f_k(\mbf{x}_i)\leq f_k(\mbf{x}_j)~\forall k \in \{1,\dotsc,n_f\}$ and
$\mbf{f}(\mbf{x}_i)\neq \mbf{f}(\mbf{x}_j)$. A solution $\mbf{x}^*
\in\Omega$ is considered \textit{Pareto-optimal} if there exists no
other solution $\mbf{x} \in\Omega$ such that $\mbf{f}(\mbf{x}) \prec
\mbf{f}(\mbf{x}^*)$. The set of all Pareto-optimal solutions is known
as the \textit{Pareto-optimal set}, defined as $\mathcal{PS}=
\left\{\mbf{x}^*\in\Omega \mid \nexists \mbf{x}\in\Omega:
\mbf{f}(\mbf{x}) \prec \mbf{f}(\mbf{x}^*)\right\}$. The image of this
set is referred to as the
\textit{Pareto-optimal front}, $\mathcal{PF} = \{ \mbf{f}(\mbf{x}^*)
\mid \mbf{x}^* \in \mathcal{PS} \}$.


A widely used way to solve MOP using classical optimization methods is
to represent the MOP as an arbitrary number of scalar optimization
problems, which are built through aggregation functions such as the
Weighted Sum (WS) or Weighted Tchebycheff
(WT)~\citep{book.Miettinen1999} approaches. Each scalar optimization
problem, generated by the aggregation function and a given weight
vector, leads to problem in which the global optimum coincides with a
particular Pareto-optimal solution of the original MOP. In this way,
an estimate of the Pareto-optimal set can be obtained by solving a set
of scalar aggregation functions.  However, simply performing an
\textit{independent} optimization of these multiple scalar problems
tends to result in difficulties for generating an adequate
approximation of the Pareto set
\citep{book.Miettinen1999,book.Deb2001}, particularly when a
well-spread sample of the Pareto-optimal front is desired.

\cite{book.Deb2001} and \cite{book.CoelloCoello2007} note that the
original MOP described in \eqref{eq:mop} can be solved
through a multiobjective evolutionary algorithm (MOEA), which is a
population-based approach that attempts to converge to an approximation of 
the Pareto-optimal set in a single run. This feature
enables a continuous exchange of information between the estimated
solutions, which is useful to promote a proper approximation of the
Pareto-optimal set. Among the algorithms commonly employed for the solution of 
continuous MOPs, we focus here on the class of MOEAs based on 
the explicit decomposition of the multiobjective optimization problem, 
which are briefly introduced below.



\subsection{Multiobjective evolutionary algorithms based on decomposition}
\label{sec:moea-d:gendesc}

Multiobjective evolutionary algorithms based on decomposition
(MOEA/Ds), originally proposed by \cite{journal.Zhang2007}, combine
features of both MOEA approaches and classical scalarization approaches 
for tackling MOPs. In general, a MOEA/D decomposes a MOP, as defined in
\eqref{eq:mop}, into a finite number of scalar optimization
\textit{subproblems}. Each subproblem is defined by a
\textit{weight vector}, which is used in an \textit{scalar aggregation
  function} to calculate the utility value of any solution for that
particular subproblem.

This set of subproblems is solved in parallel by iterating over a set
of candidate solutions, commonly called the \textit{population}. Aggregation 
functions and weight vectors are chosen so that (i) an
optimal solution to a given subproblem is also Pareto-optimal for the
original MOP; and (ii) the optimal solutions to the set of subproblems
are well distributed in the space of objectives. It is thus expected
that solving the subproblems provides a reasonable approximation of
the Pareto-optimal front, regarding both convergence and diversity criteria.

Each subproblem has one \textit{incumbent solution} that is directly
associated to it, i.e., the population size is equal to the number of
subproblems. At each iteration, one new \textit{candidate solution} is
generated for each subproblem, by the application of a sequence of
\textit{variation operators} to the existing population. This set of
new solutions is compared against each incumbent solution based on
their utility values on the respective subproblem. The best solution
for each subproblem is maintained as its incumbent solution for the
next iteration, following rules defined by a specific \textit{update 
strategy}.

When generating new candidate solutions or comparing their performance
against incumbent ones, the algorithm defines a \textit{neighborhood}
for each subproblem which limits the exchange of information between
candidate solutions. This neighborhood provides a certain locality to
the variation operators and update strategy, aiming to regulate the
convergence speed and global exploration abilities of the algorithm.

Among the features that motivate the application of a MOEA/D to the solution 
of a MOP, \cite{journal.Li2009} highlight a few that stand out in terms of their
usefulness. First, it is generally simpler to handle objective value
comparisons and, to a certain extent, diversity maintenance in a
MOEA/D than in other MOEAs.
This means that the MOEA/D is frequently able to return a set of
reasonably well-spread (in the space of objectives) solutions, even
when the number of subproblems is small or the number of objectives is
high. Scaling techniques for attenuating the effects of objective
functions with vastly different ranges are also easily incorporated
into the MOEA/D structure, as are constraint-handling techniques.


Since its introduction, the MOEA/D framework has been the target of several
investigations, mainly aimed at: i) improving its performance; ii)
overcoming limitations of its components; and iii) adapting it for
different classes of problems. For instance, studies on decomposition-based MOEAs have been carried
out to investigate new decomposition approaches
\citep{journal.LiDebZhang2014, journal.TanJiaoLi2012, journal.Qi2014,
  journal.Giagkiozis2014}, aggregation functions
\citep{inproc.Wang2013, inproc.Sato2014, inproc.Ishibuchi2010},
objective scaling strategies \citep{journal.DebJain2014,
  journal.Singh2011}, neighborhood assignment methods
\citep{inproc.Ishibuchi2013,journal.Li2014, journal.Li2015}, variation
operators \citep{journal.Li2009, journal.TanJiaoLi2012,
  journal.Li2014b}, and selection operators \citep{journal.Jiang2016}.
Additionally, decomposition-based MOEAs have been investigated for
constrained MOPs \citep{journal.DebJain2014, journal.Cheng2016},
many-objective optimization problems (MaOPs)
\citep{journal.Asafuddoula2014, journal.LiDebZhang2014}, and
incorporation of decision-maker preferences
\citep{inproc.Mohammadi2012,inproc.Gong2011,inproc.Pilat2015}.
A recent, comprehensive survey of MOEAs based on decomposition has
been organized by \cite{journal.Trivedi2016}.

Despite their specificities, we can characterize these 
different MOEA/D instantiations by defining the
following component classes in the algorithm:

\begin{itemize}
\item The \textit{decomposition strategy}, which determines how the weight
  vectors are calculated and, consequently, how the MOP gets
  decomposed.
\item The \textit{aggregation function}, which uses the weight vectors to
  generate single-objective subproblems to be solved.
\item The \textit{objective scaling strategy}, which defines how differences
  in the range of values of the objective functions are treated.
\item The \textit{neighborhood assignment strategy}, which determines the
  neighborhood relations among the subproblems. This strategy defines
  the locality of the exchange of information between candidate
  solutions when applying variation operators and updating strategies.
\item A \textit{Variation Stack}, composed of one or more
  \textit{variation operators}, which generates new candidate
  solutions from the existing ones. Our general definition of a
  variation operator is a function that modifies a candidate solution
  in the space of decision variables, based on information about the
  problem structure and/or the current and past states of the
  population as a whole. Notice that this general definition also
  encompasses \textit{repair operators} and \textit{local search
    operators} as special cases of variation.
\item The \textit{update strategy}, which determines the candidate
  solutions to be maintained or discarded after each iteration, based on
  their utility values for specific subproblems and on their
  neighborhood relations.
\item The \textit{constraint handling method}, which defines how to treat
  points that violate problem constraints.
\item The \textit{termination criteria}, which determines when the algorithm
  stops the search and returns the set of solutions found.
\end{itemize}

Based on this decomposition of the algorithm into its individual
components, it is possible to define a common framework from which
specific MOEA/D variants can be instantiated. This framework is
detailed in Section \ref{sec:moea-d}.

\section{MOEA/D implementations}
\label{sec:moead-rev}
Several implementations of the original MOEA/D and some of its
variants for continuous MOPs are available online, mainly in \proglang{Java}, \proglang{C++}
and \proglang{MATLAB}.\footnote{The \textit{MOEA/D Homepage} conveniently
  aggregates these resources in a single list, available
  at\\\url{https://sites.google.com/view/moead/home}.} All
implementations mentioned here have their source codes readily
available for download from the links listed in the references.

Table \ref{tab:moead-codes} summarizes these implementations. It is
important to notice that there is no implementation native to \proglang{R}, and that while jMetal \citep{inproc.Nebro2015} and the MOEA
Framework \citep{code.Hadka2017a} represent object-oriented frameworks
with some component-oriented design, no MOEA/D program was found to
provide the fully modular implementation of the MOEA/D as proposed in
the \pkg{MOEADr} package.

\begin{table}[htbp]
	\centering
	\label{tab:moead-codes}
	\begin{tabular}{llll}
		\hline
		Algorithm & Language(s) & Framework & Author(s)  \\
		\hline
		\multirow{2}{*}{Original MOEA/D}&\proglang{C++}   &\multirow{2}{*}{--}&\multirow{2}{*}{\cite{code.Li2006a}}\\
		&\proglang{MATLAB}   & & \\
		\hline
		Original MOEA/D&\proglang{Java}   & -- & \cite{code.Liu2006a} \\
		\hline
		MOEA/D-DE&\proglang{C++}   & -- & \cite{code.Li2007a} \\
		\hline
		\multirow{2}{*}{MOEA/D-DRA}&\proglang{C++}   &\multirow{2}{*}{--} & \multirow{2}{*}{\cite{code.Zhang2009a}}\\
		&\proglang{MATLAB}   & & \\
		\hline
		Original MOEA/D &\multirow{4}{*}{\proglang{Java}} &\multirow{4}{*}{\pkg{jMetal}}& \multirow{4}{*}{\cite{inproc.Nebro2015}}\\
		MOEA/D-DE & & &\\
		MOEA/D-DRA & & &\\
		MOEA/D-STM & & &\\
		\hline
		Original MOEA/D & \proglang{Java} &\pkg{MOEA Framework} &\cite{code.Hadka2017a}\\
		\hline
	\end{tabular}
	\caption{MOEA/D implementations for continuous MOPs.}
\end{table}

The \pkg{MOEADr} package was motivated by a perceived need to
facilitate not only the application of existing MOEA/D variants but
also the development and investigation of new components, as well as
the fast reproduction of newly published innovations based on a
library of existing components, with minimal need for
reimplementation.

Finally, it is worth mentioning that while there are no specific
MOEA/D implementations native to \proglang{R}, a few packages
implementing different, general-purpose algorithms for multiobjective
optimization are available. Table \ref{tab:moea-r} lists those which
are readily available on CRAN.\footnote{Even though most are not
  listed under the \textit{Optimization and Mathematical Programming
    Task View} (\url{https://CRAN.R-project.org/view=Optimization}) as
  of the writing of this article.}

\begin{table}[htbp]
	\centering
	\label{tab:moea-r}
	\begin{tabular}{lll}
		\hline
		Package & Algorithm(s) & Author(s)  \\
		\hline
		\pkg{goalprog} & Goal programming & \cite{code.Novomestky2008}\\
		\hline
		\pkg{NSGA2r} & NSGA-II & \cite{code.Tsou2013a}\\
		\hline
		\pkg{mopsocd} & MOPSO &\cite{code.Naval2013a}\\
		\hline
		\pkg{mco} & NSGA-II & \cite{code.Mersmann2014a}\\
		\hline
		\pkg{GPareto} & Gaussian Process & \cite{code.Binois2016a}\\
		\hline
		\multirow{3}{*}{\pkg{moko}} & HEGO & \multirow{3}{*}{\cite{code.Passos2016}}\\
		& MEGO &\\
		& VMPF &\\
		\hline
		\multirow{3}{*}{\pkg{ecr}} & NSGA-II &\multirow{3}{*}{\cite{code.Bossek2017}}\\
		& SMS-EMOA & \\
		& AS-EMOA &\\
		\hline
	\end{tabular}
	\caption{Multiobjective optimization packages native to R.}
\end{table}

Most of those packages offer a closed, individual algorithm, with the
exception of Jakob Bossek's \pkg{ecr} package \citep{code.Bossek2017}, which offers a modular
approach for instantiating evolutionary algorithms for both single and
multiobjective optimization. For the latter class, the algorithmic
structure defined by this package lends itself easily for the
implementation of dominance and indicator-based approaches, but not
necessarily for decomposition-based algorithms such the MOEA/D. This
is also the case of another recent initiative to define a common
framework for defining multiobjective evolutionary algorithms
\citep{journal.Bezerra2016}. To the authors' knowledge, there is no
\proglang{R} package implementing specific MOEA/D algorithms, nor any
component-based implementation that allows an easy instantiation of
multiobjective evolutionary algorithms based on decomposition.

\section[The MOEADr package]{The \pkg{MOEADr} package}
\label{sec:moea-d}
\label{sec:moea-d:genstruct}

The \pkg{MOEADr} package is an \proglang{R} implementation of
multiobjective evolutionary algorithms based on decomposition,
including many MOEA/D variants found in the literature. Our goals in
this package are three-fold:
\begin{itemize}
\item To provide a component-wise perspective on the MOEA/D family of
  algorithms, where each different algorithm exists as a configuration
  of common components;
\item To be easily extensible, so that researchers and users can implement
  their own components, facilitating both applied uses and scientific inquiries in this field;
\item To include a representative section of the existing literature
  in MOEA/D variants;
\end{itemize}

To achieve these goals, the implementation was guided by the following design decisions:

\begin{enumerate}
\item For the sake of uniformity in the implementation of each
  component, we define the main variables of the MOEA/D: the solution
  set, neighborhood set, subproblem weight set, subproblem utility
  value set and violation value set as matrices;
\item Algorithms in the MOEA/D family are broken down into common
  components, and each component is recast as an operator on the
  matrices defined above;
\item Each individual component avoids, to the maximum degree
  possible, to produce and rely on side effects beyond the
  explicit manipulation of the main variables above;
\item Each component is implemented as a separate \proglang{R} script file with a
  fixed naming scheme. The choice of components is specified as a
  parameter to the main function call;
\item In particular, we define a \emph{Variation Stack}, which is a
  list of variation components to be used by the MOEA/D in order. The
  choice of variation operators shows a very large diversity in MOEA/D
  literature, and using a variation stack allows for a uniform
  description of the many existing configurations;
\end{enumerate}

The general framework of the MOEA/D, as implemented in our package, is
presented in Algorithm~\ref{alg:moea-d}. The user defines an instance
of this framework by choosing a specific component for each of the
roles described in section~\ref{sec:moea-d:gendesc}, and one or more
variation components to compose the variation stack $\mathcal{V}$.

\begin{algorithm*}
	\caption{Component-wise MOEA/D structure}\label{alg:moea-d}
	\begin{algorithmic}[1]
		\Require Objective functions $\mbf{f}(\cdot)$; Constraint
		functions $\mbf{g}(\cdot)$; Component-specific input parameters;    
		\State $t\leftarrow 0$
		\State $run\leftarrow \mbox{TRUE}$
		\State Generate initial population $\mbf{X}^{(t)}$ by random
		sampling from $\Omega$.
		\State Generate weights $\mbs{\Lambda}$ \Comment{Decomposition
			strategies (Sec. \ref{sec:decomp})}
		\While{$run$}
		\State Define or update neighborhoods $\mbs{B}$
		\Comment{Neighborhood assignment strategies
			(Sec. \ref{sec:neighbor})}
		\State Copy incumbent solution set $\mbf{X}^{(t)}$ into $\mbf{X}^{\prime~(t)}$
		\For{each variation operator $v \in \mathcal{V}$}
		\State $\mbf{X}^{\prime~(t)} \leftarrow v(\mbf{X}^{\prime~(t)})$
		\Comment{Variation operators (Sec.~\ref{sec:variation})}
		\EndFor
		\State Evaluate solutions in $\mbf{X}^{(t)}$ and
		$\mbf{X}^{\prime~(t)}$ \Comment{Aggregation functions (Sec. \ref{sec:aggfun})}
		\Statex \Comment{Constraint handling (Sec. \ref{sec:chandling})}
		\State Define next population $\mbf{X}^{(t+1)}$\Comment{Update
			strategies (Sec.~\ref{sec:update})}
		\State Update $run$ flag\Comment{Stop criteria
			(Sec.~\ref{sec:stopcrit})}
		\State $t\leftarrow t+1$
		\EndWhile
		\vspace{.10cm}
		\State\Return
		$\mbf{X^{(t)}};~\mbf{f}\left(\mbf{X}^{(t)}\right)$
	\end{algorithmic}
\end{algorithm*}

In the following sections we detail each of these component classes,
presenting a formal definition of their roles as well as relevant
examples. The list of components described in this paper (and
currently available in the \pkg{MOEADr} package) is summarized in
Table~\ref{tab:moeadr}.

\renewcommand{\arraystretch}{1.4}
\begin{table*}[p]
  \label{tab:moeadr}
  \centering
  \begin{tabular}{|c|c|c|c|}
    \hline
    \textbf{Component role} & \textbf{Name} & \textbf{User parameters} & \textbf{Section}\\
    \hline
    \multirow{3}{*}{Decomposition method}& 
	SLD& $h\in\mathbb{Z}_{>0}$& \ref{sec:decomp:sld}\\
	\cline{2-4}
    & MSLD & $\mbf{h}\in\mathbb{Z}^{K}_{>0}$; $\mbs{\tau}\in\left(0,1\right]^{K}$ & \ref{sec:decomp:msld}\\
    \cline{2-4}
    & Uniform& $N\in\mathbb{Z}_{>0}$& \ref{sec:decomp:uniform}\\
    \hline
    \multirow{5}{*}{Scalar aggregation function}& 
    WS& -- &\ref{sec:aggfun:ws} \\
    \cline{2-4}
    & WT& -- & \ref{sec:aggfun:wt}\\
    \cline{2-4}
    & AWT& -- & \ref{sec:aggfun:awt}\\
    \cline{2-4}
    & PBI& $\theta^{pbi}\in\mathbb{R}_{>0}$& \ref{sec:aggfun:pbi}\\
    \cline{2-4}
    & iPBI& $\theta^{ipbi}\in\mathbb{R}_{>0}$& \ref{sec:aggfun:ipbi}\\
    \hline
Objective scaling& 
	-- & $type\in\left\{none;~simple\right\}$ &\ref{sec:aggfun}\\
    \hline
\multirow{2}{*}{Neighborhood assignment}& 
	\multirow{2}{*}{--} & $type\in\left\{by~\mbs{\lambda_i};~by~\mbf{x}_i^{(t)} \right\}$ &\ref{sec:neighbor}\\
	\cline{3-4}
	& & $\delta_p\in\left[0,1\right]$& \ref{sec:variation:probs}\\
        \hline
        
\multirow{9}{*}{Variation operators}& 
SBX recombination & $\eta_{\mathtt{X}}\in\mathbb{R}_{>0}$; $p_{\mathtt{X}}\in\left[0,1\right]$ & \ref{sec:variation:ops:sbx}\\
	\cline{2-4}
	& Polynomial mutation & $\eta_{\mathtt{M}}\in\mathbb{R}_{>0}$; $p_{\mathtt{M}}\in\left[0,1\right]$& \ref{sec:variation:ops:polymut}\\
	\cline{2-4}
	& \multirow{2}{*}{Differential mutation}& $\phi\in\mathbb{R}_{>0}$& \multirow{2}{*}{\ref{sec:variation:ops:diffmut}}\\
	& & $basis \in\left\{rand;~mean;~wgi\right\}$& \\
	\cline{2-4}
	& Binomial recombination&$\rho\in\left[0,1\right]$ & \ref{sec:variation:ops:binrec}\\
	\cline{2-4}
	& Truncation& --& \ref{sec:variation:ops:repair}\\
	\cline{2-4}
	& \multirow{3}{*}{Local search} &$type \in\left\{tpqa;~dvls\right\}$ & \multirow{4}{*}{\ref{sec:variation:ls}}\\
	& & $\tau_{ls}\in\mathbb{Z}_{>0}$; $\gamma_{ls}\in\left[0,1\right]$ & \\
	& & $\epsilon\in\mathbb{R}_{>0}$ (if $type = tpqa$)& \\
    \hline
\multirow{3}{*}{Update strategy}& 
	Standard& --& \ref{sec:update:snr}\\
	\cline{2-4}
	& Restricted& $n_r\in\mathbb{Z}_{>0}$& \ref{sec:update:rnr}\\
	\cline{2-4}
	& Best &$n_r\in\mathbb{Z}_{>0}$; $T_r\in\mathbb{Z}_{>0}$ & \ref{sec:update:bsr}\\
    \hline
\multirow{3}{*}{Constraint handling}&
	Penalty functions& $\beta_v\in\mathbb{R}_{>0}$& \ref{sec:chandling:penalty}\\
	\cline{2-4}
	& \multirow{2}{*}{VBR}& $type \in \left\{ts;~sr;~vt\right\}$& \multirow{2}{*}{\ref{sec:chandling:vbr}}\\
	& & $p_f\in\left[0,1\right]$ (if $type = sr$) & \\
	\hline
\multirow{3}{*}{Termination criteria}&
	Evaluations& $max_{eval}\in\mathbb{Z}_{>0}$& \multirow{3}{*}{\ref{sec:stopcrit}}\\
	\cline{2-3}
	& Iterations& $max_{iter}\in\mathbb{Z}_{>0}$& \\
	\cline{2-3}
	& Time& $max_{time}\in\mathbb{R}_{>0}$& \\
    \hline
    \end{tabular} 
    \caption{Components currently available in the MOEADr package.}
\end{table*}

\section[MOEA/D components available in the MOEADr package]{MOEA/D components available in the \pkg{MOEADr} package}
\label{sec:components}

In this section we describe in detail each of the components included
in the \pkg{MOEADr} package, version~2.1. We present a formal
definition of their roles as well as relevant examples, both from the
specific MOEA/D literature and the wider body of knowledge on
multiobjective evolutionary algorithms in general. The complete list
of roles and specific components is summarized in
Table~\ref{tab:moeadr}.

For each component, the different notations of existing works were
recast to a standard mathematical notation, to prevent confusion and
ambiguities that would inevitably arise if we tried to follow the many
different nomenclatures from the literature. Whenever possible, our
notation employs vector and matrix operations to describe the modules,
in an attempt to highlight the mathematical structure of each
component, as well as differences and similarities among variants.

Special care was taken to guarantee the modularity of the definitions
provided in the following sections, so that each component is
independent from design choices made for the others.  This
characteristic allows the free exchange of components while
guaranteeing the correct flow of the MOEA/D, at the cost of some
implementation overhead. This also simplifies the use of automated
algorithm assembly and tuning methods, as well as efforts for
replicating and testing algorithms from the literature.

To describe the components, let the following common variables be
defined: $n_f$ is the number of objective functions that compose the
MOP. $N\in\mathbb{Z}_{> 0}$ is the number of subproblems, which is
either a user-defined parameter, or calculated internally by the
decomposition strategy.  $\mbf{X}^{(t)} = \left\{\mbf{x}_i^{(t)}\mid
i=1,\dotsc,N\right\}$ denotes the set of incumbent solutions for the
subproblems at iteration $t$. With some abuse of notation,
$\mbf{X}^{(t)}$ will also denote a matrix with each row $i$ defined by
$\mbf{x}_i^{(t)}$.\footnote{Throughout this paper the distinction
  between the set and matrix interpretations will always be clear from
  the context. All other quantities defined as matrices will also be
  sometimes be treated as sets of vectors.}
$\mbs{\Lambda}\in\mathbb{R}_{\geq 0}^{N\times n_f}$ is the matrix of
subproblem weights calculated by the decomposition strategy (Section
\ref{sec:decomp}), and $\mbf{B}\in\mathbb{Z}_{>0}^{N \times T}$ is the
matrix of neighborhood relations calculated by the neighborhood
assignment strategy (Section \ref{sec:neighbor}), with $T$ denoting
the neighborhood size.
      
\subsection{Decomposition strategies}
\label{sec:decomp}

A decomposition strategy is responsible for generating the weight
vectors that characterize the set of scalar subproblems in a
MOEA/D. The number of subproblems can be either explicitly defined, or
calculated indirectly by the decomposition strategy based on
user-defined parameters.

Methods for generating aggregation weight vectors in the structure of
MOEA/D are usually borrowed from the theory of design and modeling in
experiments with mixtures \citep{journal.Chan2000}. Among the main
designs in this area, the simplex-lattice design
\citep{journal.Scheffe1958} has been the most commonly adopted in the
context of decomposition-based MOEAs. This approach, as well as a
multiple-layer variant are presented below. A third approach, based on
uniform designs, is also described.

The following definitions hold for all strategies discussed in this
section: let $\mbs{\Lambda}\in\mathbb{R}_{\geq 0}^{N\times n_f}$ be a
matrix of non-negative values. Also, let each row of $\mbs{\Lambda}$
be a vector summing to unity, $\norm{\mbs{\lambda}_i}_1 = 1$. Each row
$\mbs{\lambda}_i$ can be interpreted as the \textit{weight vector}
defining the $i$-th subproblem, and the matrix $\mbs{\Lambda}$ is
referred to as the \textit{weight matrix}. Since, given a scalar
aggregation function (Sec. \ref{sec:aggfun}), each weight vector
defines a unique subproblem, $\mbs{\lambda}_i$ will also be used to
refer to the $i$-th subproblem in the following sections.

In the \pkg{MOEADr} package, a list of available decomposition
strategies can be generated using the function
\texttt{get\_decomposition\_methods()}.

\subsubsection{Simplex-lattice design (SLD)}
\label{sec:decomp:sld}
In the simplex-lattice design
\citep{journal.Chan2000,journal.Zhang2007} the user provides a
parameter $h\in\mathbb{Z}_{>0}$ that defines both the number of
subproblems and the values of the weights. In this method, elements of
the weight matrix can only assume $h+1$ distinct values
\begin{equation}
  \label{eq:SLD}
  \lambda_{i,j} \in \bigg\{0,\frac{1}{h},\frac{2}{h},\dotsc,1\bigg\},~ \forall~ i,j.
\end{equation}

The simplex-lattice design generates $N$ weight vectors by using all
combinations (with repetition) of $n_f$ values taken from the set of
values defined in \eqref{eq:SLD}, which results in
\begin{equation}
  \label{eq:SLD-N}
  N = \binom{h+n_f-1}{n_f-1}
\end{equation}
subproblems defined when this strategy is used. In this way,
a general $(h, n_f)$-simplex-lattice can be used to represent $N$
weight vectors on the objective domain. For instance, for an arbitrary
three-objective problem $(n_f = 3)$, a value $h=18$ yields $N
= \binom{20}{2} = 190$ distinct weight vectors.

\subsubsection{Multiple-layer simplex-lattice design (MSLD)}\label{sec:decomp:msld}
To obtain a reasonable distribution of weight vectors within the $n_f$-dimensional
simplex using the simplex-lattice design it is necessary that $h \geq
n_f$ \citep{journal.DebJain2014}. While this condition is generally
harmless for MOPs with few objectives, a large number of weight
vectors is generated for high-dimensional objective domains, even in
the limit condition $h = n_f$.
On the other hand, making $h<n_f$ results in
weight vectors sparsely distributed only at the boundary of the
simplex, which jeopardizes the exploration abilities of the algorithm.

To address this issue, a multiple-layer approach can be devised. This
strategy generates $k$ subsets of weight vectors by generalizing the
user parameter $h$ to a vector of positive integers $\mbf{h} =
\left(h_1,\dotsc, h_K\right),~h_k\in\mathbb{Z}_{>0}~\forall k$. Each
constant $h_k$ is used to calculate $N_k$ weight vectors according to
the SLD method and, within each subset, the vectors are scaled down by
a factor $\tau_k\in\left(0,1\right]$ using a simple coordinate
  transformation
\begin{equation}
  \mbs{\lambda}^\prime_{k_i} = \tau_k\mbs{\lambda}_{k_i} + (1-\tau_k)/n_f
\end{equation}

with $\mbs{\lambda}_{k_i}$ being the $i$-th vector in the $k$-th
subset. Each subset must be associated with a unique user-defined
value of $\tau_k$.\footnote{$\tau_k = 1$ keeps the original coordinate
  system, while $\tau_k \rightarrow 0$ performs a maximum contraction
  of the weight vectors towards $\lambda_{i,j} = 1/n_f,~\forall~i,j$.}
The total number of subproblems defined in this method is given by $N
= \sum_{k=1}^KN_k$, and the final weight matrix $\mbs{\Lambda}$ is
composed by the scaled weight vectors from all layers.

This method ultimately amounts to generating weight vectors in layers,
with each layer corresponding to a different-sized simplex in the
space of objectives. This approach trades the loss of exploration
ability by the increased number of user-defined parameters (it
requires the definition of $2K+1$ parameter values). Common sense
seems to suggest $\tau_k = k/K$ as a reasonable heuristic for defining
the scaling factors, but there is currently neither empirical nor
theoretical support for this choice.

Finally, it must be remarked that this approach to generating the
weight vectors generalizes the method of
\cite{journal.LiDebZhang2014}, which was defined specifically for $K =
2$.

\subsubsection{Uniform design (UD)}
\label{sec:decomp:uniform}

The uniform design method represents an alternative approach to
generate the weight vectors. Its use was proposed by
\cite{journal.TanJiaoLi2012}, with the stated objectives of improving
the distribution of the weight vectors and providing greater control
over the number of subproblems.

The UD method for calculating the weight vectors can be
described as follows. First, let $H_N$ be defined as the set
\begin{equation}
H_N = \left\{h \in \mathbb{Z}_{>0}\mid h < N~\wedge~gcd\left(h,N\right) = 1\right\},
\end{equation}

where $gcd\left(\cdot\right)$ returns the greatest common divisor of
two integers, and $N\in\mathbb{Z}_{>0}$ is a user-defined value that
determines the number of subproblems. Let $\mbf{h} =
\left(h_1,\dotsc,h_{n_f-1}\right)$ be a vector composed of $n_f-1$
mutually exclusive elements of $H_N$. Any such vector can define a
matrix $\mbf{U}_N(\mbf{h})\in\left(\mathbb{Z}_{>0}\right)^{N \times
  (n_f-1)}$ with elements calculated as $u_{ij} = (i h_j) ~mod~
N$. Denoting the set of all possible $\mbf{h}$ vectors defined from
$H_N$ as $\mathcal{P}\left(H_N\right)$, the next step is to identify
the vector that results in the $\mbf{U}_N(\mbf{h})$ matrix with the
lowest $CD_2$ discrepancy
\begin{equation}
  \overline{\mbf{h}} =
  \underset{\mbf{h}\in\mathcal{P}\left(H_N\right)}{\arg\min}
  CD_2\left(\mbf{U}_N(\mbf{h})\right),
\end{equation}

where $CD_2(\cdot) $ is the centered $L_2$-discrepancy of a matrix
\citep{book.Fang2003,journal.TanJiaoLi2012}, calculated as
\begin{equation}
\label{eq:CD2}
\begin{split}
   & CD_2(\mbf{U}_N(\mbf{h})) = \bigg(\dfrac{13}{12}\bigg)^{(n_f-1)} - \dfrac{2}{N} \sum_{i=1}^{N} \prod_{j=1}^{n_f-1} \bigg(1 + \dfrac{\lvert u_{i,j} - 0.5 \rvert - \lvert u_{i,j} - 0.5 \rvert^2}{2} \bigg) +\\
  &\dfrac{1}{N^2} \sum_{i=1}^{N}\sum_{k=1}^{N} \prod_{j=1}^{n_f-1}\bigg(1 + \frac{\lvert u_{i,j} - 0.5 \rvert + \lvert u_{k,j} - 0.5 \rvert}{2} - \frac{\lvert u_{i,j} - u_{k,j} \rvert}{2} \bigg).\\
\end{split}
\end{equation}

If we define $\mbf{\overline{U}}_N =
\left(\mbf{U}_N(\overline{\mbf{h}}) - 0.5\right)/N$, then the elements
of the weight matrix $\mbs{\Lambda}$ are returned by the
transformation

\begin{equation}
  \begin{cases}
    \displaystyle \lambda_{i,j} = \Bigg(1-\overline{u}_{i,j}^{\big(\frac{1}{n_f-j}\big)}\Bigg) \prod_{k=1}^{j-1} \overline{u}_{i,k}^{\big(\frac{1}{n_f-k}\big)} & \mbox{, if } j < n_f \\
    \displaystyle \lambda_{i,n_f} = \prod_{k=1}^{n_f-1} \overline{u}_{i,k}^{\big(\frac{1}{n_f-k}\big)},&~\\
  \end{cases}
\end{equation}

with the guarantee that $\mbs{\Lambda}$ conforms with the properties
stated at the beginning of section \ref{sec:decomp}.
          
\subsection{Scalar aggregation functions}
\label{sec:aggfun}

The scalar aggregation function is used to calculate the utility value
of candidate solutions for each subproblem. This is done by specific
functions of the objective function values
$\mbf{f}\left(\mbf{x}\right)$ and the weight matrix $\mbs{\Lambda}$.

In the \pkg{MOEADr} package, a list of available aggregation functions
can be obtained using function \texttt{get\_scalarization\_methods()}.

\subsubsection{Weighted sum (WS)}
\label{sec:aggfun:ws}
This technique performs a convex combination of the objective
values, resulting in scalar problems of the form
\begin{equation}
  \mbox{min } f^{ws}(\mbf{x} \mid \mbs{\lambda}_i, \mbf{z}) =
  \mbs{\lambda}_i\left(\mbf{f}\left(\mbf{x}\right)-\mbf{z}\right)^\top,
  \mbox{subject to: } \mbf{x} \in \Omega
\end{equation}

where $\mbf{z} = (z_1,\dotsc,z_{n_f})$ is a reference vector with the
property that $\forall j~z_j \leq \min \{f_j(\mbf{x}) \mid \mbf{x} \in
\mathcal{PS}\}$. A diverse set of Pareto-optimal solutions can be
achieved by using different weight vectors $\mbs{\lambda}_i$. However,
due to the convex nature of this operation, only convex sections of
Pareto fronts can be approximated using this strategy
\citep{book.Miettinen1999}.

It is important to highlight here that obtaining a good estimate of
$\mbf{z}$ is by itself a computationally-intensive effort. In many
cases, it is common practice to iteratively estimate this reference
point from the set of all points visited up to a given iteration,
$\mathcal{X}^{(t^\prime)} \triangleq
\bigcup_{t=1}^{(t^\prime)}\mbf{X}^{(t)}$. In these cases the elements
of the estimated reference vector, $\hat{\mbf{z}}^{(t)}$, are defined
as
\begin{equation*}
  \hat{z}^{(t^\prime)}_j =
  \min_{\mbf{x}\in\mathcal{X}^{(t^\prime)}}f_j\left(\mbf{x}\right),~~j=1,\dotsc,n_f.
\end{equation*}

This estimated vector is updated at each iteration and is frequently
employed instead of a fixed $\mbf{z}$ in the scalar aggregation
functions used with the MOEA/D.

\subsubsection{Weighted Tchebycheff (WT)}
\label{sec:aggfun:wt}
In this approach, the scalar optimization problems are defined as
\begin{equation}
  \label{eq:wtch}
    \mbox{min } f^{wt}(\mbf{x} \mid \mbs{\lambda}_i, \mbf{z}) =
    \norm{\mbs{\lambda}_i\odot\left(\mbf{f}\left(\mbf{x}\right)-\mbf{z}\right)}_{\infty},
    \mbox{ subject to: } \mbf{x} \in \Omega
\end{equation}

where $\odot$ denotes the Hadamard product, and
$\norm{\cdot}_{\infty}$ is the Tchebycheff norm. Unlike the weighted
sum approach, the weighted Tchebycheff approach is not sensitive to
the convexity of the Pareto front \citep{book.Miettinen1999}. However,
this approach offers poor diversity control, as the solution of
\eqref{eq:wtch} for equally-spaced weight vectors do not necessarily
translate to a well-spread approximation of the Pareto front
\citep{journal.Qi2014,inproc.Wang2013}.

\subsubsection{Penalty-based boundary intersection (PBI)}
\label{sec:aggfun:pbi}
The PBI aggregation function \citep{journal.Zhang2007} is an extension
of an earlier approach known as normal boundary intersection
\citep{journal.DasDennis1998}. The scalar optimization problems defined
by the PBI strategy are given as
\begin{equation}
    \mbox{min } f^{pbi}(\mbf{x} \mid \mbs{\lambda}_i, \mbf{z}) = d_{i,1}
    + \theta^{pbi} d_{i,2}, \mbox{ subject to: } \mbf{x} \in \Omega
\end{equation}
with
\begin{equation*}
    d_{i,1} = \frac{\lvert (\mbf{f}(\mbf{x})-\mbf{z})
      \mbs{\lambda}_i^\top \rvert}{\norm{\mbs{\lambda}_i}_2};
    d_{i,2} =
    \norm{\mbf{f}(\mbf{x})-\left(\mbf{z}+\frac{d_{i,1}\mbs{\lambda}_i}{\norm{\mbs{\lambda}_i}_2}\right)}_2,
\end{equation*}

where $\norm{\cdot}_2$ denotes the Euclidean norm, and $\theta^{pbi}
\in \mathbb{R}_{\geq 0}$ is a user-defined penalty parameter. Notice
that $d_{i,1}$ is related to the convergence of $\mbf{f}(\mbf{x}_i)$
towards the Pareto-optimal front, whereas the minimization of
$d_{i,2}$ provides a way to control solution diversity. This
aggregation function enables the definition of a trade-off between
convergence and diversity (in the space of objectives) by an \textit{a
  priori} adjustment of $\theta^{pbi}$, which directly influences the
performance of the MOEA/D.

\subsubsection{Adjusted weighted Tchebycheff (AWT)}
\label{sec:aggfun:awt}
An alternative that attempts to address the poor diversity control of
the weighted Tchebycheff approach is the adjusted (or transformed)
Tchebycheff scalarization function
\citep{journal.Qi2014,inproc.Wang2013}. This method defines the scalar
subproblems as:
\begin{equation}
  \label{eq:atch}
  \begin{split}
    &\mbox{min } f^{awt}(\mbf{x} \mid \mbs{\lambda}_i, \mbf{z}) = \norm{\mbs{\rho}_i\odot\left(\mbf{f}\left(\mbf{x}\right)-\mbf{z}\right)}_{\infty}\\
    &\mbox{subject to: } \mbf{x} \in \Omega~,
  \end{split}
\end{equation}

with the elements of the vector $\mbs{\rho}_i$ defined as
\begin{equation*}
  \rho_{i,j} = \frac{\left(\lambda_{i,j} +
    \epsilon\right)^{-1}}{\sum_{j=1}^{n_f} \left(\lambda_{i,j} +
    \epsilon\right)^{-1}}~~, j = 1,\dotsc,n_f,
\end{equation*}

where $\epsilon$ is a small constant added to prevent divisions by
zero.\footnote{\cite{inproc.Wang2013} set this value as $\epsilon =
  0.0001$.}  Notice that the only difference between \eqref{eq:wtch}
and \eqref{eq:atch} is the substitution of the weight vector
$\mbs{\lambda}_i$ by its respective ``normalized inverse''
$\mbs{\rho}_i$. Besides minimizing the distance between
$\mbf{f}\left(\mbf{x}\right)$ and $\mbf{z}$, this transformation
promotes the distance minimization between an objective vector
$\mbf{f}(\mbf{x})$ and its corresponding vector $\mbs{\rho}_i$
\citep{journal.Qi2014}. In this aspect, the adjusted weighted
Tchebycheff strategy presents an idea similar to that of the PBI, but
without any additional parameters.

\subsubsection{Inverted PBI (iPBI)}
\label{sec:aggfun:ipbi}

While the PBI approach defines its search based on an ideal reference
point $\mbf{z}$ which represents an (estimated) ideal solution, the
inverted PBI function uses a \textit{nadir} reference point
$\widetilde{\mbf{z}}$, defined as $\forall k~\widetilde{z}_k \geq \max
\{f_k(\mbf{x}) \mid \mbf{x} \in \mathcal{PS} \}$
\citep{inproc.Sato2014,journal.Sato2015}.\footnote{Similarly to the
  reference point $\mbf{z}$, $\widetilde{\mbf{z}}$ can also be
  iteratively estimated by the algorithm. The procedure is analogous
  to the one described earlier for the estimation of the ideal point.}
This method works by defining the scalar minimization\footnote{The
  usual expression of the inverted PBI strategy defines a
  \textit{maximization} problem, but in this paper we express it as
  the equivalent minimization problem for the sake of
  standardization.} problems
\begin{equation}
  \begin{split}
    &\mbox{min } f^{ipbi}(\mbf{x} \mid \mbs{\lambda}_i, \widetilde{\mbf{z}}) = \theta^{ipbi} d_{i,2} - d_{i,1}\\
    &~\mbox{subject to: } \mbf{x} \in \Omega
  \end{split}
\end{equation}

with
%
\begin{equation*}
  \begin{split}
    d_{i,1} &= \frac{\lvert \left( \widetilde{\mbf{z}} - \mbf{f}(\mbf{x})\right) \mbs{\lambda}_i^\top \rvert}{\norm{\mbs{\lambda}_i}_2}\\
    d_{i,2} &= \norm{\left(\widetilde{\mbf{z}}- \mbf{f}(\mbf{x}) - d_{i,1} \frac{\mbs{\lambda}_i}{\norm{\mbs{\lambda}_i}_2}\right)}_2,
  \end{split}
\end{equation*}
%
%

where, similarly to the PBI approach, $\theta^{ipbi}
\in\mathbb{R}_{\geq 0}$ is a user-defined parameter that controls the
balance between $d_{i,1}$ (convergence) and $d_{i,2}$ (diversity).

\subsubsection*{Scaling of the objective domain}
\label{sec:aggfun:scaling}

Since the range of objective functions in MOPs can present arbitrarily
large differences, an appropriate scaling of the objective domain is
sometimes used to improve the performance of the MOEA/D. When
performed, this scaling happens prior to the calculation of the scalar
aggregation function, so that scaled function values
$\bar{f}_i(\cdot)$ replace $f_i(\cdot)$ in the calculations.

Let $\mbf{z}\in\mathbb{R}^{n_f}:z_i\leq f_i(\mbf{x}_k^{(t)})~\forall
k=1,\dotsc,N$, and $\widetilde{\mbf{z}}
\in\mathbb{R}^{n_f}:\widetilde{z}_i\geq f_i(\mbf{x}_k^{(t)})~\forall
k=1,\dotsc,N$ be estimates of the ideal and nadir objective vectors at
a given iteration $t$. A straightforward way to standardize the
objective domain \citep{book.Miettinen1999,journal.Zhang2007} is to
replace each function value $f_i(\mbf{x})$ by 
\begin{equation}
  \bar{f}_i(\mbf{x}) = \frac{f_i(\mbf{x}) - z_i}{\widetilde{z}_i -
    z_i}, ~\forall i,
\end{equation}

which guarantees that $\bar{f}_i(\mbf{x}) \in\left[0,1\right]~\forall
i$.\footnote{While alternative standardization strategies can be
  employed \citep{journal.DebJain2014,journal.Singh2011}, the general
  idea remains the same.}

          
\subsection{Neighborhood assignment function}
\label{sec:neighbor}

The neighborhood assignment function generates a matrix $\mbs{B}$
defining the neighborhood relationships between subproblems. 
In general, neighborhood relations among the
subproblems are used for defining restrictions to the exchange of
information among candidate solutions when applying the variation
operators, as well as for regulating the replacement of points at the
end of every iteration \citep{inproc.Ishibuchi2013}.

Neighborhood relations can be defined either in the space of
decision variables, $\Omega$, or in the space of objectives 
$\mbf{f}\left(\Omega\right)$. The more usual
case is the definition of neighborhoods based on the distances between
weight vectors in the space of objectives
\citep{journal.Zhang2007,journal.Li2009}. Let
$\mbf{M}\in\mathbb{R}_{\geq 0}^{N\times N}$ be the symmetric matrix defined by
taking all pairwise Euclidean distances between weight vectors, i.e.,
with elements given by
\begin{equation}
  m_{i,j}= \norm{\mbs{\lambda}_i - \mbs{\lambda}_j}_2.
\end{equation}
Let $\mbf{m}_i$ denote the $i$-th row of $\mbf{M}$. For the $i$-th
subproblem, its neighborhood vector $\mbf{b}_i\in\mathbb{Z}_{>0}^{T}$
consists of the indices of the $T$ smallest elements of $\mbf{m}_i$,
with $T\in\mathbb{Z}_{>0}$ a user-defined parameter.\footnote{Notice
  that neighborhood vector $\mbf{b}_i$ will always contain the index
  $i$ (since $m_{i,i} = 0$) plus additional $T-1$ subproblem indices
  $j\neq i$.} Since weight vectors are usually kept constant
throughout the optimization process (with some exceptions, see
\cite{journal.Qi2014}), the neighborhoods have to be determined only
once in this approach, and remain fixed throughout the execution of
the algorithm.

An alternative approach uses Euclidean distances between incumbent
solutions in the space of decision variables \citep{inproc.Chiang2011}
to define the neighborhood relations among subproblems. In this case
the distance matrix $\mbf{M}$ is defined by the distances between all
vector pairs $\mbf{x}^{(t)}_i,\mbf{x}^{(t)}_j~\in\mbf{X}^{(t)}$ and
\begin{equation}
  m_{i,j} = \norm{\mbf{x}^{(t)}_i - \mbf{x}^{(t)}_j}_2,
\end{equation}
with $\mbf{x}^{(t)}_i\in\Omega$ being the incumbent solution to the
$i$-th subproblem at iteration $t$. A neighborhood vector $\mbf{b}_i$
is then composed by the indices of the $T$ subproblems whose incumbent
solutions are closest, in the space of decision variables, to the one
associated with $\mbs{\lambda}_i$. As the incumbent solutions change
across iterations, neighborhood relations must be updated whenever a
new incumbent solution is determined for any subproblem. While this
results in increased computational cost ($N(N-1)/2$ distance
calculations per iteration), this neighborhood definition may
contribute to the algorithm performance in problems for which solution
similarity in $\Omega$ does not correspond to small distances in
$\mbf{f}\left(\Omega\right)$ \citep{inproc.Chiang2011}.
 
\subsection{Variation stack}
\label{sec:variation}

As in any evolutionary algorithm, variation operators in the MOEA/D
generate new candidate solutions based on information about points
visited by the algorithm up to a given iteration. The standard MOEA/D \citep{journal.Zhang2007} employs two variation operators, namely Simulated Binary recombination (SBX) \citep{journal.DebBeyer2001}
followed by Polynomial Mutation \citep{incol.Deb1999}, a combination
still widely used in the literature
\citep{journal.Asafuddoula2014,journal.LiDebZhang2014}. Another very
successful MOEA/D version known as MOEA/D-DE
\citep{journal.Li2009,journal.TanJiaoLi2012} employs Differential
Mutation and Binomial Recombination \citep{journal.Storn1997}, followed
by Polynomial Mutation.

While these two sets of variation operators are arguably the most
widespread in the literature, any combination of variation operators
can, at least in principle, be used to drive the search mechanism of
the MOEA/D, as long as they are sequentially compatible. This includes
the possibility of employing multiple combinations of recombination
and mutation variants -- applied either sequentially or
probabilistically -- as well as the use of local search operators.
Therefore, Algorithm~\ref{alg:moea-d} employs a user-defined set of
variation operators which are applied sequentially to a population
matrix $\mbf{X}^{\prime(t)}$, copied from the incumbent solution matrix
$\mbf{X}^{(t)}$ prior to the application of the variation operators.

In the following descriptions all operators (with the exception of the
\textit{repair operators} described later) are
defined assuming that the decision variables are contained in the
interval $\left[0,1\right]$, which greatly simplifies many
calculations. Whenever a MOP is defined with
decision variables bound to different limits, it is assumed that the
variables are properly scaled prior to the optimization.

Also included in this discussion of variation operators are
\emph{Local Search Operators}, which execute a limited search of the
solution space focused on the region close to a particular
solution. Because they act as an operator that directly modifies the
solution set, in this framework we group them together with other
variation operators.

The variation operators available in the \pkg{MOEADr} package can be
listed using functions \texttt{get\_variation\_operators()} and
\texttt{get\_localsearch\_methods()}.

\subsubsection*{Definition of sampling probabilities for variation}
\label{sec:variation:probs}

Prior to the application of the variation stack, the neighborhood
effects (Sec. \ref{sec:neighbor}) must be determined. For each
subproblem, a vector $\mbf{p}_i =
\left\{p_{i,1},p_{i,2},\dotsc,p_{i,N}\right\}$ is defined as
containing the probabilities of sampling the candidate solutions
associated with each subproblem when variation operators are applied
for the $i$-th subproblem.

In a general form, sampling probabilities are defined by the
neighborhood vectors $\mbf{b}_i$ and by a user-defined parameter
$\delta_p\in\left[0,1\right]$, which represents the total probability
of sampling from the specific neighborhood $\mbf{b}_i$. Probabilities
$p_{i,j}$ are defined as
\begin{equation}
\label{eq:varpool}
p_{i,j} = \begin{cases}
\delta_p / T&~\mbox{, if}~j\in\mbf{b}_i\\
\left(1-\delta_p\right) / \left(N - T\right)&~\mbox{, otherwise.}
\end{cases}
\end{equation}

The standard MOEA/D \citep{journal.Zhang2007} samples exclusively from
the neighborhood of each subproblem ($\delta_p = 1$ and, consequently,
$p_{i,j} = 0~\forall j\notin\mbf{b}_i$). This setting intensifies local
exploration, at the cost of a loss of solution diversity, which may
compromise the effective exploration of the design space in later
iterations \citep{journal.Li2009}. Less restrictive approaches have
been used in the literature \citep{journal.Li2009,inproc.Chiang2011},
defining $\delta_p<1$, albeit usually at relatively high values. In these approaches,
while each subproblem maintains a strong bias towards using
information from subproblems indexed by $\mbf{b}_i$, access to the other
candidate solutions remains possible, enabling the
generation of a more diverse set of points by the variation operators.

\subsubsection{SBX recombination}
\label{sec:variation:ops:sbx}
Let $\eta_{\mathtt{X}}\in\mathbb{R}_{>0}$ be a user-defined parameter,
and $\mbf{u}_i\in [0,1]^{n_v}$ be a vector of uniformly distributed
random values. Also, let $\mbs{\beta}_i\in\mathbb{R}^{n_v}$ be a
vector with elements defined as
\begin{equation}
  \label{eq:recombination-sbx-beta}
  \mbf{\beta}_{i,j} =
  \begin{cases}
    (2u_{i,j})^{\frac{1}{\eta_{\mathtt{X}} + 1}} & \mbox{if } u_{i,j}  \leq  0.5\\
    [2(1 - u_{i,j})]^{\frac{1}{\eta_{\mathtt{X}} + 1}} & \mbox{otherwise.}
  \end{cases}
\end{equation}

Let $\mbf{x}_{a_i}^{\prime(t)}\in\mathbb{R}^{n_v}$ and
$\mbf{x}_{b_i}^{\prime(t)}\in\mathbb{R}^{n_v}$ be two vectors sampled from
$\mbf{X}^{\prime(t)}$ according to the sampling probabilities defined for
the $i$-th subproblem \eqref{eq:varpool}. For each subproblem, SBX
recombination
\citep{journal.DebBeyer2001,journal.Zhang2007} produces
a new candidate solution according to
\begin{equation}
  \label{eq:recombination-sbx}
  \mbf{\widetilde{x}}^{\prime(t)}_i = \begin{cases}
  \dfrac{\left(1+\mbs{\beta}_i\right)\odot\mbf{x}^{\prime(t)}_{a_i} + \left(1-\mbs{\beta}_i\right)\odot\mbf{x}^{\prime(t)}_{b_i}}{2} &,~\mbox{if}~u_i\leq p_{\mathtt{X}}\\
  \mbf{x}^{\prime(t)}_i&,~\mbox{otherwise,}
  \end{cases}
\end{equation}

where $u_i\in [0,1]$ is a uniformly distributed random value, and
$p_{\mathtt{X}}\in [0,1]$ is a user-defined parameter. After this
operator is applied for all $i\in\left\{1,\dotsc,N\right\}$, each row
of the population matrix $\mbf{X}^{\prime(t)}$ is updated with its
corresponding point
$\mbf{\widetilde{x}}^{\prime(t)}_i$.\footnote{Despite being much more
  compact than the usual definitions of SBX recombination
  \citep{journal.DebBeyer2001,journal.Zhang2007}, this one is
  equivalent provided that the stated scaling of the variables to the
  interval $x_{i,j}\in[0,1],~\forall i,j$ is maintained.}

\subsubsection{Polynomial mutation}
\label{sec:variation:ops:polymut}

Let $\eta_{\mathtt{M}}\in\mathbb{R}_{>0}$ and
$p_m\in\left[0,1\right]$ be user-defined parameters, and $\mbf{u}_i\in
[0,1]^{n_v}$ be a vector of uniformly distributed random values. For a
given candidate solution $\mbf{x}_i^{\prime(t)}\in\mbf{X}^{\prime(t)}$, let
$\mbs{\beta}_i\in\mathbb{R}^{n_v}$ be a vector defined by
\begin{equation}
  \label{eq:mutation-polynomial-beta1}
  \beta_{i,j} =
  \left[2u_{i,j} + \left(1-2u_{i,j}\right)\left(1 - x_{i,j}^{\prime(t)}\right)^{\eta^{\prime}}\right]^{\frac{1}{\eta^{\prime}}} - 1
\end{equation}
if $u_{i,j}\leq 0.5$, or
\begin{equation}
  \label{eq:mutation-polynomial-beta2}
  \beta_{i,j} =1 - \left[2\left(1-u_{i,j}\right) + \left(2u_{i,j} - 1\right)\left(x_{i,j}^{\prime(t)}\right)^{\eta^{\prime}}\right]^{\frac{1}{\eta^{\prime}}}
\end{equation}
otherwise, with $\eta{\prime} = \eta_{\mathtt{M}} + 1$. Let also
$\mbf{v}_i\in\left\{0,1\right\}^{n_v}$ be a vector with elements
sampled independently from a Bernoulli distribution with probability
parameter $p_{\mathtt{M}}$. For each subproblem, the polynomial
mutation \citep{incol.Deb1999,journal.Zhang2007} generates a candidate
solution according to
\begin{equation}
  \label{eq:mutation-polynomial}
  \mbf{\widetilde{x}}^{\prime(t)}_i = (1 - \mbf{v}_i)\odot \mbf{x}_i^{\prime(t)} + \mbf{v}_i\odot\left(\mbf{x}_i^{\prime(t)} + \mbs{\beta}_i \right).
\end{equation}

After this operator is applied for all
$i\in\left\{1,\dotsc,N\right\}$,
$\mbf{X}^{\prime(t)}$ is updated with the new points
$\mbf{\widetilde{x}}^{\prime(t)}_i$.

\subsubsection{Differential mutation}
\label{sec:variation:ops:diffmut}

Let $\phi\in\mathbb{R}_{\neq 0}$ be a user-defined
parameter\footnote{$\phi$ can be set either as a
  constant value, or defined randomly for each application of this
  operator. In the latter, it is common to independently sample
  $\phi\in\left(0,1\right]$ for each
operation, according to a uniform distribution \citep{inproc.LiZhou2014}.}, and
  $\mbf{x}_{a_i}^{\prime(t)},~\mbf{x}_{b_i}^{\prime(t)}\in\mathbb{R}^{n_v}$
  be two mutually exclusive vectors sampled from $\mbf{X}^{\prime(t)}$
  according to the sampling probabilities defined in
  \eqref{eq:varpool}. For the $i$-th subproblem the differential
  mutation operator \citep{book.Price2005} generates a new candidate
  solution as
\begin{equation}
\label{eq:differential-mutation}
\mbf{\widetilde{x}}^{\prime(t)}_i = \mathbf{x}_{i,\text{basis}}^{(t)} + \phi\left(\mbf{x}_{a_i}^{\prime(t)} - \mbf{x}_{b_i}^{\prime(t)}\right),
\end{equation}

where $\mathbf{x}_{i,\text{basis}}^{(t)}\in\mathbb{R}^{n_v}$ is a
basis vector, which can be generated in several ways. The most common
strategy used with the MOEA/D \citep{journal.Li2009} is to randomly
sample a vector from $\mbf{X}^{\prime(t)}$ (mutually exclusive with
$\mbf{x}_{a_i}^{\prime(t)}$ and $\mbf{x}_{b_i}^{\prime(t)}$) according to the
sampling probabilities defined in \eqref{eq:varpool}.

Two possible alternatives for the basis vector in the context of
MOEA/D are also suggested here. For a given subproblem $i$, let the
points $\mbf{x}^{\prime(t)}_{b_{i,k}},~b_{i,k}\in\mbf{b}_i$ be ordered
in decreasing order of utility (i.e., in increasing order of
aggregation function value). One possibility is to use the mean point
of the neighborhood
\begin{equation}
\label{eq:diffmut-mean}
\mathbf{x}_{i,\text{basis}}^{(t)} = \frac{1}{T}\sum_{k=1}^{T}\mbf{x}^{\prime(t)}_{b_{i,k}},
\end{equation}

while the second is to use a weighted global intermediate
point, which is an adaptation of the basis vector commonly used in
evolution strategies \citep{journal.Arnold2006}
\begin{equation}
\label{eq:diffmut-wgi1}
\mathbf{x}_{i,\text{basis}}^{(t)} = \sum_{k=1}^{T}w_k\mbf{x}^{\prime(t)}_{b_{i,k}}
\end{equation}

with weights given by
\begin{equation}
\label{eq:diffmut-wgi2}
\begin{split}
w_k&=\frac{w'_k}{\sum_{k=1}^{T} w'_k}\\
w'_k &= \ln( T + 0.5 ) - \ln( k ).
\end{split}
\end{equation}

Further alternative definitions of the differential mutation operator
in the MOEA/D context can be found in the literature
\citep{inproc.LiZhou2014}. Regardless of the specifics of each
definition, this operator is known to be invariant to rotation
\citep{book.Price2005,inproc.LiZhou2014}. After its application,
$\mbf{X}^{\prime(t)}$ is updated with the points
$\mbf{\widetilde{x}}^{\prime(t)}_i$.

\subsubsection{Binomial recombination}
\label{sec:variation:ops:binrec}

Let $\rho\in\left[0,1\right]$ be a user-defined parameter,
$k_i\in\{1,\dotsc,n_v\}, i=1\dotsc,N$ denote a set of randomly
selected integers, and $\mbf{u}_i\in [0,1]^{n_v}$ be a vector of
uniformly distributed random values. Also, recall that the incumbent solutions at iteration $t$ (unmodified by any variation operators) are
stored in the matrix $\mbf{X}^{(t)}$. This operator can then be expressed
by the sequential application of
\begin{align}
  \widetilde{x}_{i,j}^{\prime(t)} &=
  \begin{cases}
    x_{i,j}^{\prime(t)}&\mbox{if $u_{i,j}\leq\rho$}\\
    x_{i,j}^{(t)}&\mbox{otherwise,}
  \end{cases};~~j=1,\dotsc,n_v \label{eq:recombination-binomial}\\
  \text{and}\\
  \widetilde{x}_{i,k_i}^{(t)} &=
  \begin{cases}
    x_{i,k_i}^{\prime(t)}&\mbox{if}~\widetilde{\mathbf{x}}_i^{\prime(t)} = \mbf{x}_i^{(t)}\\
    \widetilde{x}_{i,k_i}^{\prime(t)}&\mbox{otherwise.}
  \end{cases} \label{eq:recombination-binomial-check}
\end{align}


As with the previous operators, after this one is applied to all
subproblems, each row of $\mbf{X}^{\prime(t)}$ is updated with its
corresponding point $\mbf{\widetilde{x}}^{\prime(t)}_i$.

\subsubsection{Repair operators}
\label{sec:variation:ops:repair}

Repair operators in evolutionary algorithms for continuous optimization usually
refer to strategies for ensuring that the \textit{box constraints},
i.e., pairs of constraints defined by \\$\left(\mathtt{x}_{j,min} -
x_{i,j} \leq 0~;~x_{i,j} - \mathtt{x}_{j,max}\leq 0\right)\forall j$,
are satisfied. While the literature tends to place repair operators in
a separate category from the variation ones, we argue that they
actually belong to the same class of operations in the MOEA/D
framework, as both are used to modify candidate
solutions according to specific rules.

While there are a number of repair
methods~\citep{journal.Hellwig2016}, we describe here only the simple
truncation repair, which can be defined as
\begin{equation}
  \label{eq:repair_trunc}
  \widetilde{x}^{\prime(t)}_{i,j} = \max\left(\mathtt{x}_{j,min},~ \min\left(\mathtt{x}_{j,max}, x_{i,j}^{\prime(t)}\right)\right), \forall i,j,
\end{equation}

where functions $\max(\cdot,\cdot)$ and $\min(\cdot,\cdot)$ return the
largest and the smallest of their arguments, respectively. After this
operation is performed, $\mbf{X}^{\prime(t)}$ is updated accordingly.

\subsubsection{Local search operators}
\label{sec:variation:ls}

Local search strategies are usually employed in the MOEA/D to
accelerate convergence. As with the repair operators, these approaches
are commonly classified as distinct from the variation operators, but
we argue that they are indeed part of the same block, for the same
reasons stated earlier.

Since these operators sometimes result in the need for additional
evaluations of candidate solutions, as well as in loss of diversity,
their application is usually regulated by a frequency parameter,
expressed either as a \textit{period of application}
$\tau_{ls}\in\mathbb{Z}_{>0}$ (i.e., the operator is applied to each
subproblem once every $\tau_{ls}$ iterations); or as a
\textit{probability of application} $\gamma_{ls}\in\left[0,1\right]$
(i.e., at every iteration the operator is applied with probability
$\gamma_{ls}$ for a given subproblem)
\citep{book.Deb2001,book.CoelloCoello2007}. Whenever any of these
conditions is true, local search is applied on the incumbent solution
of a subproblem \textit{instead} of all other variation
operators. Notice that both $\tau_{ls} = 1$ and $\gamma_{ls} = 1$ lead
to the same behavior, that is, the MOEA/D ignoring all other variation
operators and performing only local search at every iteration. Values
that are sometimes offered for these parameters in the wider MOEA
literature are $\tau_{ls} = 5$ \citep{journal.Wanner2008} and
$\gamma_{ls}\in\left[0.01,0.05\right]$
\citep{book.Deb2001,book.CoelloCoello2007}, albeit without much
theoretical or experimental justification.

As with all other variation methods, after this operation is performed
the rows of $\mbf{X}^{\prime(t)}$ are updated with the corresponding
points $\mbf{\widetilde{x}}^{\prime(t)}_i$.

\paragraph{Three-point quadratic approximation (TPQA):}
\label{sec:variation:ls:tpqa}

Let $f^{agg}_{i_k} \overset{\triangle}{=} f^{agg}(\mbf{x}^{\prime(t)}_{i_k} \mid \mbs{\lambda}_i,
\mbf{z})$ be the aggregation function value of the $k$th candidate
solution, $k \in \mbf{b}_i$, for the $i$th subproblem. Let
$\mbf{x}^{\prime(t)}_{i_1},~\mbf{x}^{\prime(t)}_{i_2},~\mbf{x}^{\prime(t)}_{i_3}\in\mathbb{R}^{n_v}$ be the three candidate solutions with the highest utility for the $i$th subproblem at iteration $t$, such
that $f^{agg}_{i_1} \leq f^{agg}_{i_2} \leq f^{agg}_{i_3}$. The three-point
quadratic approximation method \citep{journal.TanJiaoLi2012} for
performing local search in the MOEA/D can be expressed as
\begin{equation}
  \label{eq:tpqa}
  \widetilde{x}_{i,j}^{\prime(t)} =
  \begin{cases}
    x^{\prime(t)}_{i_1,j},&\mbox{if}~q_{i,j} < \epsilon\\
    \widehat{x}^{(t)}_{i,j},&\mbox{otherwise,}
  \end{cases};~~j=1,\dotsc,n_v
\end{equation}

where
\begin{equation}
  \label{eq:tpqa2}
  \begin{split}
    q_{i,j} = &\left(x_{i_2,j}^{\prime(t)}  - x_{i_3,j}^{\prime(t)} \right)f^{agg}_{i_1} + \left(x_{i_3,j}^{\prime(t)}  - x_{i_1,j}^{\prime(t)} \right)f^{agg}_{i_2} +\\ &~~\left(x_{i_1,j}^{\prime(t)}  - x_{i_2,j}^{\prime(t)} \right)f^{agg}_{i_3}
  \end{split}
\end{equation}
\begin{equation}
  \label{eq:tpqa3}
  \begin{split}
    &\widehat{x}^{(t)}_{i,j} = \left[\left(x_{i_2,j}^{\prime(t)}\right)^2 - \left(x_{i_3,j}^{\prime(t)}\right)^2\right]\frac{f^{agg}_{i_1}}{2q_{i,j}} + \left[\left(x_{i_3,j}^{\prime(t)}\right)^2 \right.\\
    &~~\left.- \left(x_{i_1,j}^{\prime(t)}\right)^2\right]\frac{f^{agg}_{i_2}}{2q_{i,j}}+ \left[\left(x_{i_1,j}^{\prime(t)}\right)^2 - \left(x_{i_2,j}^{\prime(t)}\right)^2\right]\frac{f^{agg}_{i_3}}{2q_{i,j}}
  \end{split}
\end{equation}

where $\epsilon\in\mathbb{R}_{>0}$ is a small positive
constant.\footnote{\cite{journal.TanJiaoLi2012} recommend $\epsilon = 10^{-6}$.}

\paragraph{Differential vector-based local search (DVLS):}
\label{sec:variation:ls:dvls}

Although this strategy was originally proposed for non-decomposition MOEAs \citep{journal.Chen2015},
its adaptation to the MOEA/D framework is straightforward. Let
$\mbf{x}_{a_i}^{\prime(t)},~\mbf{x}_{b_i}^{\prime(t)}\in\mbf{X}^{\prime(t)}$
be two candidate solutions, with $\left(a_i,~b_i\right)$ denoting
mutually exclusive indices sampled from the neighborhood
$\mbf{b}_i$. Two new candidate alternatives,
$\widehat{\mbf{x}}^{+}_{i}$ and $\widehat{\mbf{x}}^{-}_{i}$, are then
generated according to \eqref{eq:differential-mutation}, using
$\mathbf{x}_{i,\text{basis}}^{(t)} = \mbf{x}_{i}^{(t)}$ and $\phi = \pm \phi_{ls}$, with
$\phi_{ls}\sim\mathcal{N}\left(\mu = 0.5, \sigma = 0.1\right)$
\citep{journal.Chen2015}. This local search operator then compares these
two alternatives against the incumbent solution, and returns the point
with the best performance for the subproblem, i.e.,
\begin{equation}
  \label{eq:dvls}
  \widetilde{\mbf{x}}_i^{\prime(t)} = \underset{\mbf{x}\in\left\{\mbf{x}_{i}^{\prime(t)},~\widehat{\mbf{x}}^{+}_{i},~\widehat{\mbf{x}}^{-}_{i}\right\}}{\arg\min} f^{agg}(\mbf{x} \mid \mbs{\lambda}_i, \mbf{z}).
\end{equation}

\newpage
\subsection{Update strategies}
\label{sec:update}

Update strategies in the MOEA/D play the same role as
selection operators in general MOEAs, regulating the
substitution of incumbent solutions by those resulting from
the application of a sequence of variation operators. In the MOEA/D, this
substitution is controlled by the replacement strategy \citep{journal.Zhang2007,journal.Asafuddoula2014,inproc.WangZhang2014},
as well as by the neighborhood relations between subproblems. Three
update methods are presented next.

In what follows, let $\mbf{X}^{\prime (t)}$ denote the matrix
containing the candidate solutions generated after the application of
the variation stack, and $\mbf{X}^{(t)}$ be the matrix containing
the incumbent solutions at iteration $t$.  Also, let
$f^{agg}\left(\cdot\right)$ denote the aggregation function (see Sec. \ref{sec:aggfun}). For each subproblem $i$,
the update strategy will generate a set of candidate solutions $C_i$ (
generally composed of the incumbent solution $x_i^{(t)}$ and candidate
solutions from a given neighborhood) and select the best solution in this set as
the new incumbent solution $x_i^{(t+1)}$.

The list of update strategies available in the \pkg{MOEADr} package can be generated using \texttt{get\_update\_methods()}.

\subsubsection{Standard neighborhood replacement}
\label{sec:update:snr}
Let the candidate set for the $i$-th
subproblem be defined as
\begin{equation}
  \label{eq:rep-standard-candset}
  \mbf{C}_i^{(t)} = \mbf{x}_i^{(t)}\cup\left\{\mbf{x}^{\prime (t)}_k\bigl\lvert k\in\mbf{b}_i \right\},
\end{equation}

where $\mbf{b}_i $ is the neighborhood defined in Section \ref{sec:neighbor}. This replacement strategy \citep{journal.Zhang2007} updates the population according to
\begin{equation}
  \label{eq:rep-standard}
  \mbf{x}_i^{(t+1)} = \biggl\{\mbf{c}^{(t)}_q \bigl\lvert  \mbf{c}^{(t)}_q\in\mbf{C}_i^{(t)} \wedge
  f^{agg}\left(\mbf{c}^{(t)}_q \mid \mbs{\lambda}_i, \mbf{z}\right)\leq\underset{\mbf{c}^{(t)}_k\in\mbf{C}_i^{(t)}}{\min}f^{agg}\left(\mbf{c}^{(t)}_k \mid \mbs{\lambda}_i, \mbf{z}\right)\biggr\}.
\end{equation}

That is, the best solution to the $i$-th subproblem considering the
incumbent solution and the candidate solutions indexed in $\mbf{b}_i$.

\subsubsection{Restricted neighborhood replacement}
\label{sec:update:rnr}

In the standard replacement, a single candidate solution can replace up to $T$ (the
neighborhood size) incumbent solutions at any given iteration, which can lead to diversity
loss and premature stagnation. To overcome this
issue, \cite{journal.Li2009} proposed a limit to the number of copies that any single point $\mbf{x}_i^{\prime (t)}$
can pass on to $\mbf{X}^{(t+1)}$, defined by a user-defined
parameter $n_r\in\mathbb{Z}_{>0}\mid n_r\leq N$. This approach generalizes the standard
neighborhood replacement (which happens if $n_r = T$), and
provides a relatively simple way to limit the propagation speed of
candidate solutions in the population. The definition of good values
for $n_r$, however, may require some tuning: both $n_r = 2$
\citep{journal.Li2009} and $n_r = 0.01N$ \citep{inproc.Zhang2009} are recommended,
without much theoretical or empirical support.

\subsubsection{Best subproblem replacement}
\label{sec:update:bsr}

In the previous strategies, a candidate solution $\mbf{x}_i^{\prime
  (t)}$ is considered for replacing the incumbent solutions of its
neighboring subproblems, i.e., it can replace incumbent solutions
$\mbf{x}_{k}^{(t)}\mid k\in\mbf{b}_i$. However, it may happen that the
best available candidate solution for a given subproblem $i$ is not
among those in the neighborhood $\mbf{b}_i$. To address this issue, \cite{inproc.WangZhang2014} suggested a three-step
replacement strategy. The first step is to identify the subproblem
$k_i$ for which each new candidate solution $\mbf{x}_i^{\prime (t)}$
is most effective, i.e.,
\begin{equation}
  \label{eq:rep-bestsub-index}
  k_i=\underset{1\leq k\leq N}{\arg\min} f^{agg}(\mbf{x}_{i}^{\prime (t)}\lvert \mbs{\lambda}_{k}, \mbf{z}).
\end{equation}

The second step is to generate replacement neighborhoods
$\mbf{b}^r_{k_i}$, which are composed of the $T_r$ nearest neighbors to
the $k_i$-th subproblem (calculated as in Sec.
\ref{sec:neighbor}), where $T_r\in\mathbb{Z}_{>0}\mid T_r\leq N$ is a
user-defined parameter. 
Finally, for each
subproblem $i$, the original neighborhood $\mbf{b}_i$ is replaced by
$\mbf{b}^r_{k_i}$, and then the \textit{Restricted Neighborhood Replacement} is used for updating the population.

\subsection{Constraint handling approaches}
\label{sec:chandling}

Some possible approaches to deal with constraints in the MOP
formulation are discussed below. While most available implementations
of the MOEA/D (see Section \ref{sec:moead-rev}) provide versions that
are only capable of dealing with box constraints (using repair
operators such as truncation), dealing with general constraints in the
context of the MOEA/D is relatively straightforward, as discussed
below.

The constraint handling techniques available in the \pkg{MOEADr}
package can be consulted using \texttt{get\_constraint\_methods()}.

\subsubsection{Penalty functions}
\label{sec:chandling:penalty}

The most common approach in the MOEA community to handle constraints is
to use penalties \citep{book.Miettinen1999,journal.Coello2002}. The
idea of penalty functions is to transform a constrained optimization
problem to an unconstrained one, by adding a certain value to the
aggregation function value of a given point based on the magnitude of
constraint violation that it presents.

Assuming that the magnitude of the constraint violation is derived
from the sum of violations of all inequality and equality constraints,
its value can be defined as\footnote{Individual violation values can
  also be subject to scaling, to prevent extreme differences in the
  scale of constraint functions from dominating the attribution of the
  penalized utility value. This discussion, however, is not advanced
  further in the present work.}
\begin{equation}
  \label{eq:cviolation}
  v(\mbf{x}) = \sum\limits_{i=1}^{n_g} \max(g_i(\mbf{x}),0) + \sum\limits_{j=1}^{n_h} \max(\lvert h_j(\mbf{x}) \rvert - \epsilon,0),
\end{equation}

where $\epsilon \in \mathbb{R}_{\geq 0}$ is a small threshold
representing a tolerance for the equality constraints. The new
(penalized) aggregation function to be optimized is then obtained as
\begin{equation}
  \label{eq:penalagg}
  f_{pen}^{agg}(\mbf{x}\lvert\mbs{\lambda}_{k}, \mbf{z}) = f^{agg}(\mbf{x}\lvert\mbs{\lambda}_{k}, \mbf{z}) + \beta_{v}v(\mbf{x})~,
\end{equation}

in which $\beta_v\in\mathbb{R}_{>0}$ is a user-defined penalization
constant. For each subproblem, the penalized values are then used to
select which solution will become (or remain) the incumbent one in the
next population, $\mathbf{X}^{(t+1)}$, according to the update
strategies (Sec. \ref{sec:update}).

\subsubsection{Violation-based ranking (VBR)}
\label{sec:chandling:vbr}

This constraint handling technique generalizes a few methods available
in the literature, such as Tournament Selection (TS)
\citep{journal.Deb2000}, Stochastic Ranking (SR)
\citep{journal.Runarsson2000}, and Violation Threshold (VT)
\citep{inproc.Asafuddoula2012}. Like the Penalty Functions approach,
Violation-based Ranking uses the total magnitude of constraint
violations $v(\mbf{x})$ \eqref{eq:cviolation} to penalize unfeasible solutions, but this
penalization takes the form of ranking functions that employ different
quantities depending on the feasibility (or not) of the solutions
being compared.

In its general form, this constraint handling method uses the
following criteria when ranking the solutions:
\begin{itemize}
\item If a solution is feasible, use its $f^{agg}\left(\mbf{x}\lvert \mbs{\lambda}_{k}, \mbf{z}\right)$ value;
\item If a solution is unfeasible, check a criterion $c\left(\mbf{x}\right)$:
  \begin{itemize}
  \item If $c\left(\mbf{x}\right) = TRUE$, use its $f^{agg}\left(\mbf{x}\lvert \mbs{\lambda}_{k}, \mbf{z}\right)$ value;
  \item If $c\left(\mbf{x}\right) = FALSE$, use  its $v\left(\mbf{x}\right)$ value;
  \end{itemize}
\item Assuming that $n_{1}\leq N$ points are to be ranked using their $f^{agg}\left(\mbf{x}\lvert \mbs{\lambda}_{k}, \mbf{z}\right)$ values, VBR ranks those solutions first (rank values from $1$ to $n_{1}$), and then attributes ranks for the remaining solutions using their $v\left(\mbf{x}\right)$ values, from $n_{1}+1$ to $N$. Rank values are then used instead of $f^{agg}\left(\mbf{x}\lvert \mbs{\lambda}_{k}, \mbf{z}\right)$ whenever solutions must be compared.
\end{itemize}

As mentioned earlier, this criterion generalizes the
commonly-used constraint handling techniques Tournament Selection (TS),
Stochastic Ranking (SR), and Violation Threshold (VT). These specific methods
can be instantiated from the rules defined in Table
\ref{tab:vbr-criteria}, by choosing an appropriate function
$c\left(\mbf{x}\right)$, see Table \ref{tab:vbr-specs}. For SR, $p_f\in\left[0,1\right]$ represents a user-defined parameter, and
$u\in\left[0,1\right]$ denotes a uniformly distributed random
value. For VT, the adaptive threshold $\epsilon_{v}$ is calculated
independently for each subproblem, as
\begin{equation}
  \epsilon_{v,i} = \frac{[fs]_i}{\left(T + 1\right)^2}\sum_{k\in\mbf{b}^{\prime}_i}v\left(\mbf{x}_k\right),
\end{equation}

where $\mbf{b}_i^{\prime} = \mbf{x}_i^{(t)} \bigcup
\left\{\mbf{x}_{k}^{\prime(t)}\lvert k\in\mbf{b}_i\right\}$; $[fs]_i$
represents the number of feasible solutions in $\mbf{b}_i^{\prime}$;
and $(T+1)$ is the cardinality of $\mbf{b}^{\prime}_i$.\footnote{If
  the \textit{best subproblem replacement} is used, then
  $\mbf{b}^r_{k_i}$ is used instead of $\mbf{b}_i$, and $T_r$ instead
  of $T$.}
\renewcommand{\arraystretch}{1.1}
\begin{table}[htb]
  \centering
  \label{tab:vbr-criteria}
  \begin{tabular}{|c|c|c|}
    \hline
    \multicolumn{2}{|c|}{Situation}&  Rank using\\
    \hline
    \multicolumn{2}{|c|}{$v(\mbf{x}) = 0$}& \multirow{2}{*}{$f^{agg}\left(\mbf{x}\lvert \mbs{\lambda}_{k}, \mbf{z}\right)$}\\
    \cline{1-2}
    \multirow{2}{*}{$v(\mbf{x})> 0$}& $c\left(\mbf{x}\right) = TRUE$& \\
    \cline{2-3}
    &$c\left(\mbf{x}\right) = FALSE$ & $v\left(\mbf{x}\right)$\\
    \hline
  \end{tabular}
  \caption{Ranking criteria for the violation-based ranking.}
\end{table}
\renewcommand{\arraystretch}{1.1}
\begin{table}[htb]
  \centering
  \label{tab:vbr-specs}
  \begin{tabular}{|c|c|}
    \hline
    Method & Function $c\left(\mbf{x}\right) = $\\
    \hline
    Tournament selection \citep{journal.Deb2000} & $FALSE$\\
    Stochastic ranking \citep{journal.Runarsson2000}& $\left(u \leq p_f\right)$\\
    Violation threshold \citep{inproc.Asafuddoula2012}& $\left(v\left(\mbf{x}\right)\leq\epsilon_{v}\right)$\\
    \hline
  \end{tabular}
  \caption{Secondary criterion functions.}
\end{table}

An important point to highlight here is that the VBR method does not
necessarily guarantee that feasible solutions will always be
maintained in the population. This means that a good feasible solution
obtained previously for a scalar subproblem may be eliminated later in
the search. To prevent this loss of feasible solutions, an elitist
archiving strategy can be adopted in order to preserve the
\textit{feasible} solution with the best aggregation value found for
each sub-problem throughout the iterations. In this case, this elitist
archive should be considered as the output population of the algorithm
\citep{inproc.Ying2016}.

Finally, it is worth mentioning that this generalization of known
constraint handling techniques may serve as a template for further
developments of better methods for dealing with constraints in
population-based algorithms, both for single and multi-objective
cases. This possibility, however, falls outside the scope of the
current paper, and will not be explored further.

\subsection{Termination criteria}
\label{sec:stopcrit}

Termination criteria determine when the algorithm should stop
running and return the solutions found.
Common criteria in the evolutionary computation literature
tend to fall into two broad categories, namely: (i) exhaustion of the
algorithmic budget, in terms of execution time, number of points
visited, or number of iterations performed; and (ii) attainment of a
specific threshold value for some criterion, such as lack of
improvement, loss of population diversity, convergence to specified
quality values, or some other population statistic.\footnote{An
  interesting compilation of works on stop
  criteria for MOEAs is maintained by Luis Mart\'i at\\\url{http://lmarti.com/stopping}.}

\textit{Time-based} criteria stop the process after a user-defined
amount of time has passed since the program started running. These
criteria can be useful when comparing computationally intensive tasks
such as the construction of neighborhood or subproblem weight
matrices, but it can be sensitive to background tasks if \textit{wall
  clock} is used. In general, \textit{process time} tends to be
preferable, although it tends to add some complexity to the
implementation.

\textit{Number of iterations} criteria stop the search
immediately after the iteration counter becomes higher than a given
threshold. This tends to be useful when comparing minor
variations of a given algorithm, as long as there is no
significant differences in the computational cost per iteration.

Finally, \textit{number of evaluations} criteria terminate the
search when the total number of points visited by the algorithm
reaches a predefined threshold. This criterion is usually preferred in
situations when the larger part of the total computational cost is
incurred by utility function evaluations, in which case it is a good
practice to keep this value constant when comparing very different
algorithms. Notice that because the termination check occurs once per
iteration, the total number of evaluations may show some fluctuations
from the user-defined value.

The list of available stop criteria in the \pkg{MOEADr} package can be
consulted using function \texttt{get\_stop\_criteria()}.

\section{Usage examples}
\label{sec:experiments}

In this section we present three case studies which illustrate common
usage situations for the \pkg{MOEADr} package. The first shows
how to solve a simple MOP using the package. The second illustrates
the application of the component oriented framework with an automatic 
algorithmic assembling and tuning approach. 
Finally, the third example demonstrates how to use a
user-defined component with the package. 

These three case studies are also available as vignettes in the \pkg{MOEADr} documentation.

\subsection[Solving a simple multiobjective optimization problem using MOEADr]{Solving a simple multiobjective optimization problem using \pkg{MOEADr}}

In this case study, we use \pkg{MOEADr} to optimize a simple
2-objective problem composed of the (shifted) Sphere and Rastrigin
functions in the domain $\Omega = \left[-1,1\right]^{n_v}$, defined as
\begin{equation}
\begin{split}
\text{sphere}(\mbf{x})  &= \sum_{i=1}^{n_v} \left(x_i + 0.1i\right)^2\text{, and}\\
\text{rastrigin}(\mbf{x})  &= \sum_{i=1}^{n_v} \left[\left(x_i - 0.1i\right)^2 - 10 \text{cos}\left(2\pi \left(x_i - 0.1i\right)\right) + 10\right]
\end{split}
\end{equation}

The \proglang{R} implementation of the problem can be defined as
follows. Note that the \pkg{MOEADr} package requires the
multiobjective problem to be defined as a function that can receive a
population matrix $\mathbf{X}$, and return a matrix of objective
function values for each point.
\begin{Sinput}
R> #Define objective functions 
R> sphere <- function(X) {
+    X.shift <- X + seq_along(X) * 0.1 
+    sum(X.shift**2) }
R> rastringin <- function(X) {
+    X.shift <- X - seq_along(X) * 0.1 
+    sum((X.shift)**2 - 10 * cos(2 * pi * X.shift) + 10) }
R> #wrap objective functions in an evaluator routine
R> problem.sr <- function(X) {
+    t(apply(X, MARGIN = 1,
+    FUN = function(X) {c(sphere(X), rastringin(X))})) }
R> #assemble a problem definition list
R> problem.1 <- list(name = "problem.sr",
+    xmin = rep(-1, 30), xmax = rep(1, 30), m = 2)
\end{Sinput}

To load the package and run the problem using the original MOEA/D \citep{journal.Zhang2007}, we
use the following commands:
\begin{Sinput}
R> library("MOEADr")
R> results <- moead(problem = problem.1,
     preset = preset_moead("original"), seed = 42)
\end{Sinput}

The \code{moead()} function requires a problem definition, discussed
above, an algorithm configuration, logging parameters, and a seed.  In
this example we used an algorithm preset. The \code{preset\_moead()}
function can output a number of different presets based on
combinations found on the literature. These presets can also be
modified (either partially or as a whole), as will be shown in another
case study. \code{preset\_moead("original")} returns the original
MOEA/D configuration, as proposed by \cite{journal.Zhang2007}. Running
\code{preset\_moead()} without an argument outputs a list of available
presets.

The \code{moead()} function returns a list object of class
\code{moead}, containing the final solution set, objective
values for each solution, and other information about the optimization
process.  The \pkg{MOEADr} package uses S3 to implement versions of
\code{plot()}, \code{print()} and \code{summary()} for this object.

\begin{figure}
  \begin{center}
    \includegraphics[width=0.6\textwidth]{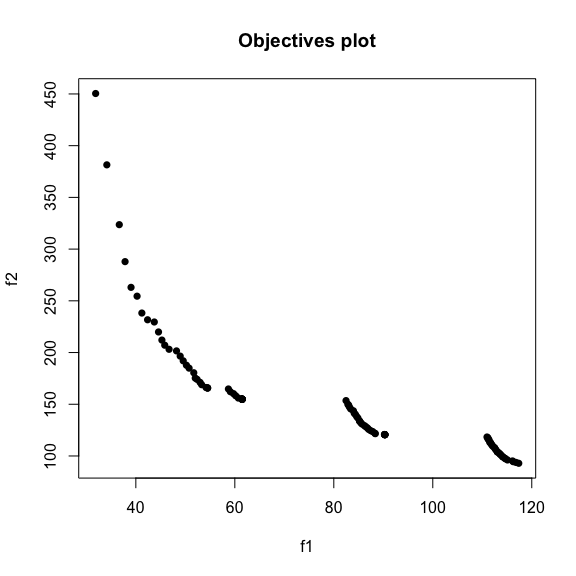}
  \end{center}
  \caption{Estimated front produced by running the code in Case Study 1.1.}
  \label{fig:case11}
\end{figure}

\code{plot()} will show the estimated Pareto front for the objectives,
as in Figure \ref{fig:case11}.  When the number of objectives is
greater than 2, a parallel coordinates plot is also produced (see
Figure \ref{fig:case12}). \code{summary()} displays information about
the number of non-dominated and feasible solution points, the
estimated ideal and nadir values, and (optionally) the inverted
generational distance (IGD) and hypervolume indicators
\citep{journal.Zitzler2003} calculated for the set of feasible,
nondominated points returned.

\begin{Sinput}
R> summary(results)
R> plot(results)
\end{Sinput}
\begin{Soutput}
#> Warning: reference point not provided:
#>    using the maximum in each dimension instead.
#> Summary of MOEA/D run
#> #====================================
#> Total function evaluations:  20100
#> Total iterations:  200
#> Population size:  100
#> Feasible points found:  100 (100% of total)
#> Nondominated points found:  100 (100% of total)
#> Estimated ideal point:  31.879 92.925
#> Estimated nadir point:  117.372 450.424
#> Estimated HV:  23972.73
#> Ref point used for HV:  117.3718 450.4242
#> #====================================

## (Plot output shown in Figure 1)  
\end{Soutput}

\subsubsection{A more complex example}

Package \pkg{smoof} \citep{smoof} provides the implementation of a large number of
single and multiobjective test functions. \pkg{MOEADr} 
provides a wrapper function \code{make\_vectorized\_smoof()} to easily
convert \pkg{smoof} functions to the format required by the
\code{moead()} function, as illustrated below for a 5-objective standard MOP benchmark problem.

\begin{Sinput}
R> library("smoof")
R> DTLZ2 <- make_vectorized_smoof(prob.name = "DTLZ2",
+    dimensions = 20, n.objectives = 5)
R> problem.dtlz2 <- list(name = "DTLZ2", xmin = rep(0, 20),
+    xmax = rep(1, 20), m = 5)
\end{Sinput}

The code below shows an example of how to modify an algorithm
preset. Because of the higher number of objectives, we want to reduce
the value of the parameter $H$ in the SLD decomposition component (see
Section~\ref{sec:decomp}) used by the preset from 100 to 8:

\begin{Sinput}
R> results.dtlz <- moead(problem = problem.dtlz2,
+    preset = preset_moead("original"),
+    decomp = list(name = "SLD", H = 8), seed = 42)
\end{Sinput}

\begin{figure}
  \begin{center}
    \includegraphics[width=0.49\textwidth]{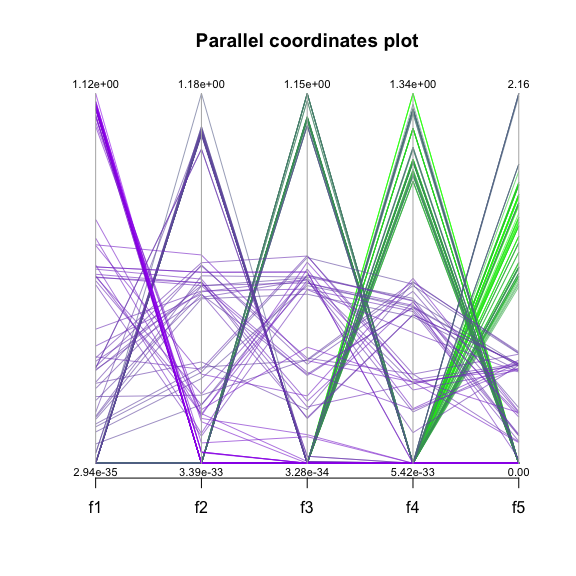}
    \includegraphics[width=0.49\textwidth]{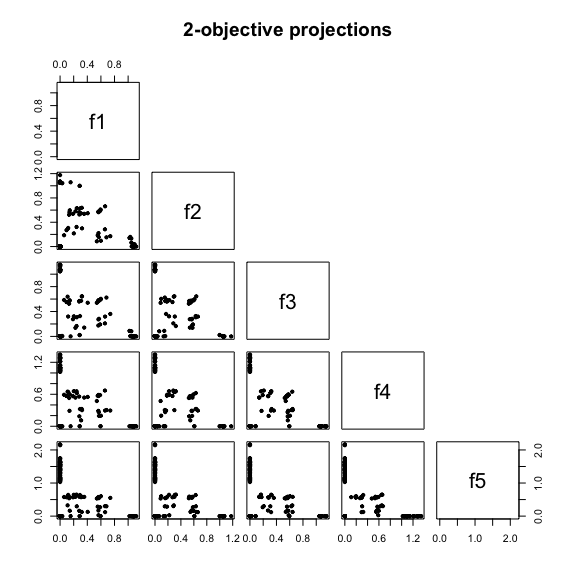}
  \end{center}
  \caption{Parallel coordinates plot and 2 objective projections produced by Case Study 1.2.}
  \label{fig:case12}
\end{figure}

As before, we see an overview of the results with \code{summary()} and
\code{plot()}. Notice that the output of  \code{plot()} is different when the number of objectives in a problem is greater than 2, as shown in Figure \ref{fig:case12}.

\begin{Sinput}
R> summary(results.dtlz)
R> plot(results.dtlz)
\end{Sinput}
\begin{Soutput}
#> Warning: reference point not provided:
#>    using the maximum in each dimension instead.
#> Summary of MOEA/D run
#> #====================================
#> Total function evaluations:  99495
#> Total iterations:  200
#> Population size:  495
#> Feasible points found: 495 (100% of total)
#> Nondominated points found:  242 (48.9% of total)
#> Estimated ideal point: 0 0 0 0 0
#> Estimated nadir point: 1.252 1.176 1.174 1.571 2.16
#> Estimated HV:  4.974365
#> Ref point used for HV: 1.252253 1.176129 1.174102 1.57124 2.160007
#> #====================================

## (Plot output in Figure 2)
\end{Soutput}

\subsection[Fine tuning algorithm configurations using MOEADr and irace]{Fine tuning algorithm configurations using \pkg{MOEADr} and \pkg{irace}}

For this example, we employ the \emph{Iterated Racing}
procedure (available in the \pkg{irace} package) \citep{journal.Ibanez2016},
to automatically assemble and fine-tune a MOEA/D configuration based
on the components available in the \pkg{MOEADr} package.

Ten unconstrained test problems from the CEC2009
competition\footnote{Functions UF1 to UF10
  \url{http://dces.essex.ac.uk/staff/qzhang/moeacompetition09.htm},
  using the standard implementation available in the \pkg{smoof}
  package \citep{smoof}.} are used, with dimensions ranging from 10 to
30. Dimension 20 was reserved for testing, while all
others were used for the training effort. To quantify the quality of
the set of solutions returned by a candidate configuration we use the
Inverted Generational Distance (IGD) \citep{journal.Zitzler2003}
indicator.  The number of subproblems was fixed as $100$ for $n_f = 2$ and
$150$ for $n_f = 3$.

We define a tuning budget of 1,000 runs for this example. The algorithm was set with the following fixed parameters: \textit{Uniform} decomposition, \textit{AWT} agreggation function, \textit{restricted} update, \textit{simple} scaling, \textit{max. evaluations = $50,000$} as a stop criterion. The variation stack was partially fixed with the sequence of operators \textit{differential mutation} (with $\phi\sim U\left(0,1\right)$), \textit{binomial recombination}, \textit{polynomial mutation} (with $p_\mathtt{M} = 1 / n_v$) and \textit{simple truncation} repair. 

Seven tunable parameters were set as follows:
\begin{itemize}
\item Neighborhood $type \in\left\{by~\mbs{\lambda}_i;~~by~\mbf{x}_i^{(t)}\right\}$;
\item Neighborhood size $T\in\left[10,40\right]$;
\item Probability of sampling from neighborhood $\delta_p\in\left(0.1, 1.0\right)$;
\item For the restricted update: $n_r\in\left[1,10\right]$;
\item Differential mutation $basis \in\left\{rand,~mean,~wgi\right\}$;
\item Binomial recombination $\rho \in\left(0,1\right)$;
\item Polynomial mutation $\eta_{\mathtt{M}}\in\left(1,100\right)$;
\end{itemize}

The code of this case study requires a large amount of boilerplate for
the \pkg{irace} package \citep{journal.Ibanez2016} and, for the sake of brevity, it is not
included in the text. Please refer to the supplementary materials for the 
full replication script.\footnote{A much larger version of this experiment is also present in 
the ``Fine tuning MOEA/D configurations using MOEADr and irace''
vignette in the \pkg{MOEADr} package.}

\begin{figure}[h]
	\centering
	\resizebox{\linewidth}{!}{\input{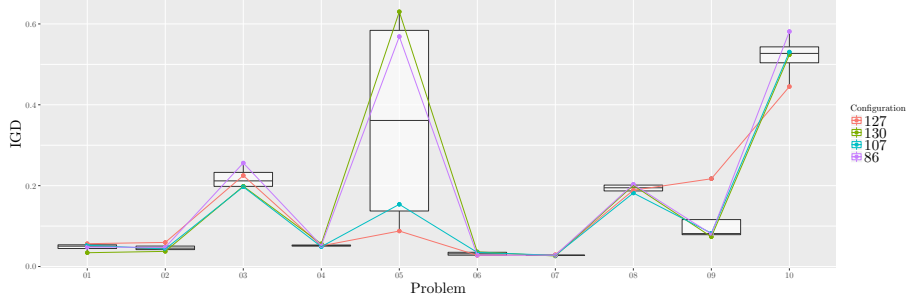}}
	\caption{IGD values for the 4 final configurations returned by the iterated racing procedure
		on the testing problems.}
	\label{fig:casestudy2}
\end{figure}

\begin{figure}
  \centering
  \resizebox{\linewidth}{!}{\input{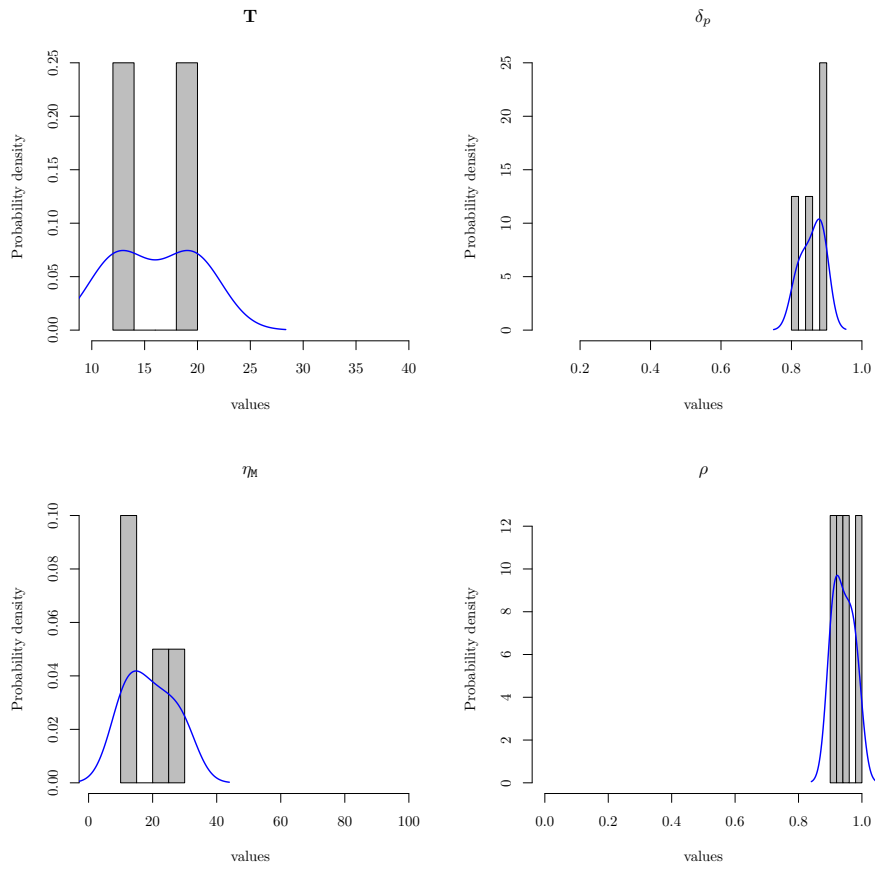}}
  \caption{Values of the numeric parameters returned by irace.}
  \label{fig:pars}
\end{figure}

Figure~\ref{fig:casestudy2} shows the IGD values achieved by the four
final configurations over the test problems. Based on these final
configurations, we assemble the consensus MOEA/D configuration, which
is described in Table \ref{tab:moead-irace}. The third column of this
table, together with Fig. \ref{fig:pars}, provides the consensus value
of each component, measured (in the table) as the rate of occurrence
of each component in the seven final configurations returned by
Iterated Racing procedure. These results suggest that the automated
assembling and tuning method reached a reasonably solid consensus, in
terms of the components used as well as the values returned for the
numeric parameters.
\renewcommand{\arraystretch}{1.1}
\begin{table}[h!]
  \label{tab:moead-irace}
  \centering
  \begin{tabular}{|c|c|c|}
    \hline
    \textbf{Component}	& \textbf{Value}	& \textbf{Consensus}\\ 
    \hline
    Decomposition 		& Uniform				& Fixed\\
    \hline
    Aggregation function & AWT				& Fixed\\
    \hline
    Objective scaling 	& simple 			& Fixed\\
    \hline
    \multirow{3}{*}{Neighborhood}& by $\mbf{x}$		& $1.00$\\
    & $T = 13$			& see Fig. \ref{fig:pars}\\
    &$\delta_p = 0.887$ & see Fig. \ref{fig:pars}\\
    \hline
    \multirow{7}{*}{Variation stack} & Differential mutation & Fixed\\
    &$basis = ``rand"$ 	 & $1.00$\\
    &$\phi\sim U(0,1)$ 	& Fixed\\
    \cline{2-3}
    & Binomial recombination	& Fixed\\
    &$\rho= 0.906$			& see Fig. \ref{fig:pars}\\
    \cline{2-3}
    & Polynomial mutation	&Fixed\\
    &$p_\mathtt{M} = 1/n_v$ 	 & Fixed\\
    &$\eta_\mathtt{M} = 10.429$			& see Fig. \ref{fig:pars}\\
    \cline{2-3}
    & Truncate					& Fixed\\
    \hline
    \multirow{3}{*}{Update}   		& Restricted				& Fixed\\
    &$n_r = 3$					&$0.75$\\
    \hline
  \end{tabular} 
  \caption{Final MOEA/D configuration returned by the iterated racing procedure.
      ``Value'' shows the best configuration, and ``Consensus''
      represents the agreement between the four final configurations.}
\end{table}

This case study illustrates how to use the \pkg{MOEADr} package to 
explore the space of possible component
configurations and parameter values, which can render improved 
algorithmic configurations and new insights into the roles of specific 
components and parameter values outside the standard values from the
literature. The example used in this paper is intended only as a 
proof-of-concept, but we highly recommend that a similar approach is
used when developing new components, in order to observe not only the
individual performance of the novel component, but also its
interaction with components which already exist in the MOEA/D
environment.
\newpage

\subsection[Adding new components to MOEADr]{Adding new components to \pkg{MOEADr}}
\label{sec:casestudy3}

In addition to a wide variety of components from the literature, the
\pkg{MOEADr} package also supports the use of user-defined components.
For this example, we will create a simple variation operator that does
not exist in the package in its current state, and compare it with the
original MOEA/D.

Consider the following ``Gaussian Mutation'' operator: given a set $X$
of solutions $\mathbf{x_i} \in X$ we add, with probability $p$, a
noise $r_{ij} \sim \mathcal{N}(\mu, \sigma)$ to each $x_{ij} \in
\mathbf{x_i} \in X$. The \proglang{R} code for this operator is as
follows:

\begin{Sinput}
R> variation_gaussmut <- function(X, mean = 0, sd = 0.1, p = 0.1, ...) {
R>   # vector of normally distributed values, 
R>   #length(R) = nrow(X) * ncol(X)
R>   R <- rnorm(length(X), mean = mean, sd = sd)
R>   # Apply random binary mask, probability = p
R>   R <- R * (runif(length(X)) <= p)
R>   # Add mutations to the solution matrix 
R>   return (X + R)
R> }
\end{Sinput}

We would like to highlight a few characteristics of the code sample
above.  First, the name of the function must be in the form
\code{variation\_[functionname]}.  The \pkg{MOEADr} package uses
function name prefixes to perform some automated functions such as
listing existing components and error checking. The list of current
function prefixes and their meaning is:
\begin{itemize}
\item \code{constraint\_}: Constraint handling components
\item \code{decomposition\_}: Decomposition functions
\item \code{ls\_}: Local search operators
\item \code{scalarization\_}: Scalarization functions
\item \code{stop\_}: Stop criteria components
\item \code{uptd\_}: Update components
\item \code{variation\_}: Variation operators
\end{itemize}

Second, the parameters in the definition of the variation operator
function must include: the solution set matrix \code{X}, any specific
parameters for the function, and finally an ellipsis argument to catch
any other inputs passed down by the main \code{moead()} function, such
as objective values and former solution sets. Extensively commented
examples of using these parameters are available in the source code
for the variation operators included in the package, such as the
Binomial Recombination operator (\code{variation\_binrec()}). Other
component classes follow similar rules, as documented in the
\textit{Writing Extensions for the MOEADr Package} vignette.

If a given function is not available in the \pkg{MOEADr} package
environment, it will search for it in the base \proglang{R}
environment.  Therefore, if you have named your component correctly,
all you need to do is add it to the appropriate parameter in the
\code{moead()} call.

For example, let us replace the variation stack of the original MOEA/D
by our Gaussian Mutation operator, followed by simple truncation, and
test it on a standard benchmark function, and use the plotting
function to perform a purely qualitative graphical comparison against
the original MOEA/D:
\begin{Sinput}
R> library("MOEADr")
R> library("smoof")
R> ZDT1 <- make_vectorized_smoof(prob.name = "ZDT1", dimensions = 30)
R> problem.zdt1 <- list(name = "ZDT1", xmin = rep(0, 30), 
+    xmax = rep(1, 30), m = 2)
R> 
R> # Variation stack, with our new operator followed by truncate
R> myvar <- list()                                   
R> myvar[[1]] <- list(name = "gaussmut", p = 0.5)    
R> myvar[[2]] <- list(name = "truncate")             
R> 
R> results.orig <- moead(problem = problem.zdt1,
+    preset = preset_moead("original"),seed = 42)
R> results.myvar <- moead(problem = problem.zdt1,
+    preset = preset_moead("original"), variation = myvar, seed= 42)
R> plot(results.orig)
R> plot(results.myvar)
\end{Sinput}

\begin{figure}[htb]
\begin{center}
  \includegraphics[width=.49\textwidth]{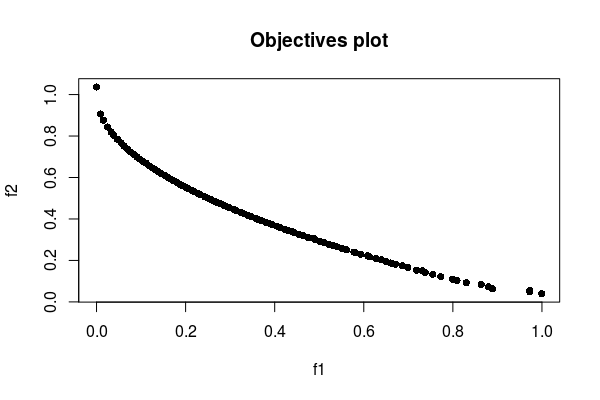}
  \includegraphics[width=.49\textwidth]{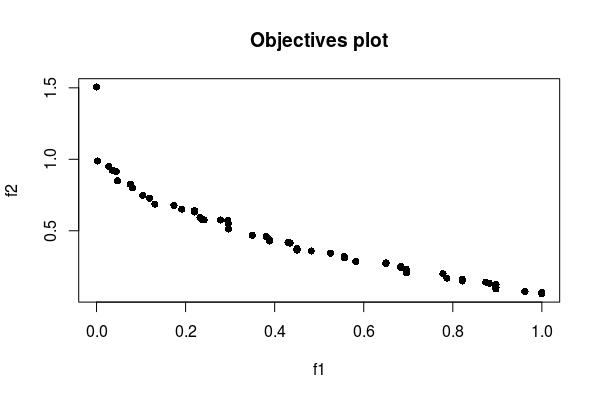}
\end{center}
\caption{Estimated Pareto-front achieved by the ``original'' MOEA/D composition
  (left) and the MOEA/D using the Gaussian mutation variation (right).}
\label{fig:cs3}
\end{figure}

Figure~\ref{fig:cs3} shows the estimated Pareto front for both the
standard MOEA/D and the MOEA/D with a Gaussian Mutation operator. From
these images it seems, rather unsurprisingly, that the example
operator is not an adequate replacement for the variation methods used
in the standard MOEA/D.

\section{Summary and conclusions}
\label{sec:conclusion}

In this article we presented the \proglang{R} package \pkg{MOEADr},
implementing a new, component-based formulation of the MOEA/D
framework. This formulation breaks down the algorithm into independent
components that can be separately replaced or configured. We described
many recent proposals in the MOEA/D literature in an unified
mathematical formulation fitting this framework.

The package allows users to easily instantiate a large variety of
works from the MOEA/D literature, which facilitates the
reproducibility of published results and the application of known
variants to problems of interest. By implementing the components
listed in Table~\ref{tab:moeadr} and allowing the definition of new
functions for any component of the algorithm, the package also allows
practitioners to quickly test new proposals and ideas.

Recently, \cite{journal.Bezerra2016} introduced a component-wise
framework for dominance- and indicator-based MOEAs (a vision that is
partially implemented in the \pkg{ecr} package by \cite{code.Bossek2017}), and described how
that framework was useful for the automated generation of
algorithms. In the present work we introduced a counterpart for
decomposition-based approaches, covering the third major branch of
multiobjective evolutionary algorithms.

Besides expanding the library of available components, current work
for improving the package includes the addition of parameter
self-adaptation and of performance-based stop criteria. The
possibility of including preference information within the
\pkg{MOEADr} framework to bias the search towards specific subsets of
the Pareto-optimal front
\citep{journal.Goulart2016,inproc.Goulart2017a}, as well as expanding
the package to deal with robust multiobjective optimization problems
\citep{inproc.Goulart2017b} are also among the improvements planned
for a future release.

\section*{Acknowledgments}
The first two authors (FC and LSB) would like to acknowledge the
financial support of Brazilian funding agencies CNPq and FAPEMIG,
which allowed them to pursue the projects that eventually gave rise to
the idea of a component-based framework and of the \pkg{MOEADr}
package.

Authors FC and CA contributed equally to testing and development,
while LSB provided expertise regarding the MOEA/D algorithm and its
many variants, as well as the initial formalization of the algorithmic
components.

\bibliographystyle{acm}
\bibliography{10_bibliography}

\end{document}